\documentclass{article}

\PassOptionsToPackage{numbers, compress}{natbib}

\usepackage[final]{neurips_2022}
\usepackage{algorithm, algpseudocode}%
\floatname{algorithm}{Algorithm}

\usepackage{amsthm}

\algtext*{EndFunction}

\def\RR{{\mathbb R}}    %r�els
\def\EE{{\mathbb E}}    % espérance
\def\11{{\mathbf 1}}    % indicatrice

       \def\cH{{\mathcal H}}       \def\cU{{\mathcal U}}        \def\cK{{\mathcal K}}       \def\cX{{\mathcal X}} \def\cY{{\mathcal Y}}  

\def\bfx{{\bf x}} \def\bfy{{\bf y}} \def\bfz{{\bf z}} 
 \def\bfK{{\bf K}}

\def\bfu{{\bf u}}

\def\bfX{{\bf X}}
\def\bff{{\bf f}}
\def\bfu{{\bf u}}
\def\bfU{{\bf U}}
\def\bfXleft{\mathbf{X}^{(\ell)}}
\def\bfXright{\mathbf{X}^{(r)}}
\def\Xright{X^{(r)}}
\def\Xleft{X^{(\ell)}}
\def\bfxleft{\mathbf{x}^{(\ell)}}
\def\bfxright{\mathbf{x}^{(r)}}
\def\Sc{{S^c}}
\def\pnu{\nu^{(p_I)}}

\def\balpha{\boldsymbol{\alpha}}

\usepackage[utf8]{inputenc} % allow utf-8 input
\usepackage[T1]{fontenc}    % use 8-bit T1 fonts
\usepackage{url}            % simple URL typesetting
\usepackage{booktabs}       % professional-quality tables
\usepackage{amsfonts}       % blackboard math symbols
\usepackage{nicefrac}       % compact symbols for 1/2, etc.
\usepackage{microtype}      % microtypography
\usepackage{xcolor}         % colors
\usepackage{amsmath}
\usepackage{hyperref}

\usepackage{cleveref}
\usepackage{graphicx}
\usepackage{subcaption}
\usepackage{lipsum}  
\usepackage{wrapfig}
\usepackage{enumitem}

\newtheorem{definition}{Definition}[section]
\newtheorem{proposition}{Proposition}[section]

\title{Explaining Preferences with Shapley Values}

\author{%
Robert Hu\thanks{Equal contribution, order decided by coinflip}\hspace{0.05cm} \thanks{Work primarily done at the University of Oxford and finished at Amazon.} \\
Amazon\\
London\\
\And
Siu Lun Chau\footnotemark[1] \\
Department of Statistics \\
University of Oxford \\
\And
Jaime Ferrando Huertas\\
Shaped\\
New York \\
\And 
Dino Sejdinovic \\
Department of Statistics \\
University of Oxford \\
}

\begin{document}
\newcommand{\shortparagraph}[1]{\textbf{#1}\quad}

\maketitle

\begin{abstract}

% key point 1: Explaining preferences are useful but underexplored.
% point 2: we propose pref shap.
% point 3: we show more meaningful explanations obtained compared to naive comparison with utiltiy based models.

While preference modelling is becoming one of the pillars of machine learning, the problem of preference explanation remains challenging and underexplored. In this paper, we propose \textsc{Pref-SHAP}, a Shapley value-based model explanation framework for pairwise comparison data. We derive the appropriate value functions for preference models and further extend the framework to model and explain \emph{context specific} information, such as the surface type in a tennis game. To demonstrate the utility of \textsc{Pref-SHAP}, we apply our method to a variety of synthetic and real-world datasets and show that richer and more insightful explanations can be obtained over the baseline.

\end{abstract}

\vspace{-0.5cm}
\section{Introduction}
\vspace{-0.3cm}

Preference learning~\cite{10.1007/978-3-540-39857-8_15} is a classical problem in machine learning, where one is interested in learning the order relations on a collection of data items.  Preference learning algorithms~\citep{bradley1952rank,thurstone1994law,chu2005preference,gonzalez2017preferential} often assume that there is a latent utility function $f: \mathcal X \mapsto \mathbb{R}$ dictating the outcome of preferences, where $\mathcal X$ denotes the domain of item covariates. An explicit feedback such as item ratings or rankings from recommender systems can be treated as noisy evaluations of $f$, whereas pairwise comparison data~(also known as duelling data) arising from, e.g., sports match outcomes~\cite{cattelan2013dynamic, chau2020spectral} can be used to implicitly infer $f$, i.e. item $\bfxleft$ is preferred over (beats) item $\bfxright$ when $f(\bfxleft) > f(\bfxright)$. As shown by \citet{kahneman1979interpretation}, humans often struggle with evaluating absolute quantities when it comes to eliciting preferences, but are broadly capable of evaluating relative differences, a core observation often exploited in preference learning. Motivated by such, this work will focus on explaining preferences inferred using duelling data.

% However, as discussed by \citet{chau2020learning}, direct feedback from $f$ is often scarce or expensive, and the quantity of implicit feedback data typically far outweighs the explicit data. Moreover, when the feedback is provided by humans,~\citet{kahneman1979interpretation} showed that we are better at evaluating relative differences than absolute quantities. Motivated by this observation, this work will focus on explaining preferences inferred using pairwise comparison data.

% why is explaining preference important

Explaining preference models is crucial when they are applied in areas such as recommendation systems~\cite{houlsby2012collaborative}, finance~\cite{bennett2022lead}, and sports science~\cite{stuart2006multiple} for the practitioner to trust, debug and understand the value of their findings~\cite{chau2021rkhs}. However, despite its importance, no prior work has studied this problem to the best of our knowledge. While one may suggest applying existing explainability tools such as \textsc{LIME}~\cite{lime}, or \textsc{SHAP}~\cite{lundberg2017unified} to a learned utility function $f$, we reason  %\textcolor{red}{(or in section 2)} 
that this approach only explains the utility but not the mechanism of eliciting preferences itself. We highlight the important differences between these two viewpoints in our numerical experiments. Moreover, the utility-based model places a strong \textit{rankability} assumption on the underlying preferences, meaning that if we define $\bfxleft \preceq \bfxright \iff f(\bfxleft) \leq f(\bfxright)$, then $\preceq$ is a total order on all the items. However, as \citet{pahikkala2010} and \citet{chau2020learning} have discussed, there are many departures from rankability in practice, e.g. we might easily see a preference of $A$ over $B$, $B$ over $C$, but $C$ over $A$ -- conforming to the \textit{rock-paper-scissors} relation. Such inconsistent preferences are under frequent study in social choice theory~\cite{sep-social-choice,gehrlein1983condorcet}, and are of wider interest in both healthcare~\cite{pref_learning_medicine} and retail~\cite{retail_pref} where data are both large and noisy. 

To move beyond the rankability assumption, we will utilise the \textit{Generalised Preferential Kernel} from \cite{chau2020learning} to model the underlying preferences, and develop \textsc{Pref-SHAP}, a novel Shapley value~\cite{shapley1953value}-based explainability toolbox, to explain the inferred preferences. Our contributions can be summarised as follows:

% \quad 1. We propose \textsc{Pref-SHAP}, a novel Shapley value-based explainability algorithm, to explain preferences.

% \quad 2. We empirically demonstrate that \textsc{Pref-SHAP} gives more informative explanations compared to the naive approach of applying \textsc{SHAP} to the utility function $f$.

% \quad 3. We release a computationally optimized implementation of \textsc{Pref-SHAP} at [github].

\begin{enumerate}
    \item We propose \textsc{Pref-SHAP}, a novel Shapley value-based explainability algorithm, to explain preferences based on duelling data.
    \item We empirically demonstrate that \textsc{Pref-SHAP} gives more informative explanations compared to the naive approach of applying \textsc{SHAP} to the inferred utility function $f$.
    \item We release a high-performant implementation of \textsc{Pref-SHAP} at~\cite{pref_shap_code}.
\end{enumerate}

\vspace{-0.4cm}
\section{Background materials}
\vspace{-0.2cm}

%\section{Background materials}
\label{sec: background}
We will first give a brief overview of preference learning and Shapley Additive Explanations~(SHAP)~\cite{lundberg2017unified}, which are the two core concepts of our contribution, \textsc{Pref-SHAP}, described in Section~\ref{sec: method}.\newline \shortparagraph{Notation} Scalars are denoted by lower case letters, while vectors and matrices are denoted by bold lower case and upper case letters, respectively. Random variables are denoted by upper case letters. $\cX \subseteq \RR^d$ denotes the item space with $d$ features and $\cY = \{-1, 1\}$ is the binary preference outcome space\footnote{Thus, we do not model `draws' in match outcomes, but the model can be straightforwardly extended to include them by specifying the appropriate likelihood function.}. We let $k: \cX\times \cX \to \RR$ be a kernel function and $\cH_k$ the corresponding reproducing kernel Hilbert space (RKHS).

\vspace{-0.2cm}
\subsection{Preference Learning}
\vspace{-0.2cm}
\label{sec: preference_learning}

In this section, we will introduce the two approaches to model preferences from duelling data, namely the \textit{utility based approach} and the more general approach from \citet{chau2020learning}. Formally, a preference feedback is denoted as \textit{duelling}, when a pair of items $(\bfxleft, \bfxright) \in \cX \times \cX$ is given to a user, and a binary outcome $y\in\cY$ telling us whether $\bfxleft$ or $\bfxright$ won the duel, is observed. In general, we observe $m$ binary preferences among $n$ items, giving the data $D = \left(\bfy, \bfXleft, \bfXright\right) = \left\{ (y_j, \bfxleft_j, \bfxright_j) \right\}_{j=1}^m$. We also use $\bfX \in \RR^{n\times d}$ to denote the full item covariate matrix.

\shortparagraph{Utility-based Preference model (UPM)}
The following likelihood model is often used~\cite{bradley1952rank, thurstone1994law, chu2005preference, gonzalez2017preferential,chau2020spectral} to model duelling feedback using a latent utility function $f$:
\begin{align}
\label{eq: simple_pref}
    p\left(y\mid \bfxleft, \bfxright \right) = \sigma\left(y\left(f\left(\bfxleft\right) - f\left(\bfxright\right)\right)\right),
\end{align}
where $\sigma$ is the logistic CDF, i.e. $\sigma(z) = (1+\exp(-z))^{-1}$. Maximum likelihood approaches are then deployed to learn the latent utility function $f$. Consequently, preferences between items can be inferred accordingly from $\bff = \{f(\bfx_i)\}_{i=1}^n$, i.e. $\bfx_i$ is on average preferred over $\bfx_j$ if $\bff_i \geq \bff_j$.

Albeit elegant, there are several drawbacks to this approach in modelling preferences. As mentioned, using a one-dimensional vector $\bff$ to derive preferences assumes that the items $\{\bfx_i\}_{i=1}^n$ are perfectly rankable, i.e. there is a total ordering on $\cX$ which the true preferences are consistent with. This is a strong assumption that often does not hold in practice. For example, it is well studied that cognitive biases often lead to inconsistent human preferences in behavioural economics~\cite{kahneman1979interpretation}. Moreover, the ranking community has also challenged this assumption by devising rankability metrics~\cite{anderson2019rankability, cameron2020graph} to test this restrictive assumption in practice. 

\shortparagraph{Generalised Preference Model (GPM)} \citet{chau2020learning} proposed to model preference directly using a more general $g: \cX \times \cX \to \RR$ that captures the preference within any pair of items, using the likelihood
\begin{align}\label{eq:gpm_likelihood}
    p\left(y \mid \bfxleft, \bfxright\right) = \sigma\left(yg\left(\bfxleft, \bfxright\right)\right).
\end{align}
We note that $g$ has to be a skew-symmetric function to ensure the natural property $p(y\mid \bfxleft, \bfxright) = 1 - p(y\mid \bfxright, \bfxleft)$. The utility based approach can be obtained as a special case of this model, i.e. by setting $g(\bfxleft, \bfxright) = f(\bfxleft) - f(\bfxright)$. We propose that when one is interested in modelling (and thus explaining) pairwise preferences, we should consider the preference function $g$ directly instead of explaining preferences based on a restrictive utility model $f$. 

% This belief has also led to several extensions of preference models~\cite{causer2005, chen2016} that do not assume rankability.

%We argue that when one is interested in modelling pairwise preferences, we should explicitly model the preference itself, rather than indirectly inferring preference from a utility model that restricts the preference structure. This belief has led to several extensions of preference models~\cite{chau2020learning, causer2005, chen2016} that do not assume rankability. In particular, \citet{chau2020learning}'s model, which is the core methodology of our contribution \textsc{Pref-SHAP}, considered the generalisation of Eq~\ref{eq: simple_pref},
% \begin{align}
%     p\left(y \mid \bfxleft, \bfxright\right) = \sigma\left(yg\left(\bfxleft, \bfxright\right)\right)
% \end{align}
% where $g:\cX \times \cX \to \RR$ is a skew-symmetric function to ensure the preferences are skew-symmetrical, a natural restriction since we need $p(y\mid \bfxleft, \bfxright) = 1 - p(y\mid \bfxright, \bfxleft)$. 
We follow \citet{chau2020learning}'s approach to model $g$ non-parametrically using kernel methods~\cite{paulsen2016introduction}. We assume $g$ as a function lives in the following RKHS of skew-symmetric functions: given kernel $k:\cX\times\cX\to\RR$ defined on the item space $\mathcal X$, the \textit{generalised preferential kernel} $k_E$ on $\mathcal X \times \mathcal X$ is constructed as follows:
\begin{align*}
\small
    k_E\left(\left(\bfxleft_i, \bfxright_i\right), \left(\bfxleft_j, \bfxright_j \right)\right) = k\left(\bfxleft_i, \bfxleft_j\right)k\left(\bfxright_i, \bfxright_j\right) - k\left(\bfxleft_i, \bfxright_j\right)k\left(\bfxright_i, \bfxleft_j\right).
\end{align*}
This kernel allows us to model the similarity across pairs of items. Moreover, if $k$ is a universal kernel~\cite{sriperumbudur2011universality}, then $k_E$ also satisfies the corresponding notion of universality, meaning that the corresponding RKHS $\cH_{k_E}$ is rich enough to approximate any bounded continuous skew-symmetric function arbitrarily well~\cite[Theorem. 1]{chau2020learning}. To infer $g\in\mathcal H_{k_E}$ using likelihood \eqref{eq:gpm_likelihood}, one simply runs kernel logistic regression with data $\bfy$ as labels and $\left(\bfXleft, \bfXright\right)$ as inputs. We will refer to this approach as the \textit{Generalised Preference Model}~(GPM).\\ \newline We emphasize that \emph{explaining} GPM allows us to specifically explain \emph{inconsistent preferences}, which in contrast to \emph{explaining rank} allows us to infer preferences even when transitivity is violated. Such insights can be of great importance in broader contexts such as decision theory \cite{Anand1987} and utility theory \cite{10.2307/1913746} where transitivity does not hold.
% and provide implementation details of the algorithm in the appendix.

\shortparagraph{Incorporating context variables.} Besides item-level covariates $\bfx\in\cX$, when there exist additional \emph{context covariates} $\bfu \in \cU \subseteq \RR^{d'}$ that describe the context in which a specific pairwise comparison is made, they can be incorporated into the kernel design as discussed in \citet[Appendix.~B]{chau2020learning}. Examples of such context covariates could be court type when a tennis match is conducted, or where a different user compares two clothing items in e-commerce.  Considering the enriched dataset  $D = \left\{ \left(y_j, \bfu_j, \bfxleft_j, \bfxright_j\right) \right\}_{j=1}^m$, we can now model the preference incorporating the context as: $p(y \mid \bfu, \bfxleft, \bfxright) = \sigma\left(g_U\left(\bfu, \bfxleft, \bfxright \right)\right)$. Now, given a kernel $k_U$ defined on the context space $\cU$, the context-specific preference function $g_U:\cU\times \cX\times \cX \to \RR$ can be learnt non-parametrically with the following kernel,
\begin{align*}
\small
    k_E^{(U)}\left(\left(\bfu_i, \bfxleft_i, \bfxright_i\right), \left(\bfu_j, \bfxleft_j, \bfxright_j \right)\right) = k_U\left(\bfu_i, \bfu_j\right) k_E\left(\left(\bfxleft_i, \bfxright_i\right), \left(\bfxleft_j, \bfxright_j \right)\right).
\end{align*}
We refer to this approach as the \textit{Context-specific Generalised Preference Model}~(C-GPM).

\vspace{-0.2cm}
\subsection{Shapley Additive Explanations~(SHAP)}
\vspace{-0.2cm}
To explain preferences, we will utilise the popular SHAP (SHapley Additive exPlanations) paradigm, which is based on the concept of Shapley values (SV). SV~\cite{shapley1953value} were originally proposed as a credit allocation scheme for a group of $d$ players in the context of cooperative games, which are characterised by a value function $\nu: [0,1]^d \to \RR$ that measures \textit{utility} of subsets of players. Formally, the Shapley value for player $j$ in game $\nu$ is defined as:
% \begin{align*}
% \label{eq: shapley}
% \small
% \phi_j(\nu) = \frac{1}{d}\sum_{S \subseteq \Omega \backslash \{j\}} {d-1 \choose |S|}^{-1} \left(\nu(S\cup j) - \nu(S)\right),
% \end{align*}
\begin{align}
\label{eq: shapley}
\small
\phi_j(\nu) = \sum_{S \subseteq \Omega \backslash \{j\}} (|S|!(d-|S|-1)!/{d!}) \left(\nu(S\cup j) - \nu(S)\right),
\end{align}
where $\Omega = \{1, ..., d\}$ is the set of players of the game. Given a value function $\nu$, the Shapley values are proven to be the only credit allocation scheme that satisfies a particular set of favourable and fair game theoretical axioms, commonly known as \textit{efficiency}, \textit{null player property}, \textit{symmetry} and \textit{additivity}~\cite{shapley1953value}. \citet{vstrumbelj2014explaining} later connect Shapley values to the field of \textit{explainable machine learning} by drawing an analogy between model fitting and cooperative game. Given a specific data point, by considering its \textit{features} as \textit{players} participating in a game that measures features' utilities, the Shapley values obtained can be treated as \emph{local feature importance scores}. Such games are typically defined through the value functions defined below.
\begin{definition}[Value functions]
Let $X$ be a random variable on $\cX\subseteq\RR^d$ and $f: \cX\to\RR$ a model from hypothesis space $\cH$. The value function $\nu:\cX\times[0,1]^d\times\cH $ is given by
\begin{align}
    \nu_{\bfx, S}(f) = \EE_{r\left(X_{S^c}\mid X_S=\bfx_S\right)}\left[f(\{X_S, X_{S^c}\})\mid X_S=\bfx_S\right]
\end{align}
where $r$ is an appropriate reference distribution, $X_S$ is the subvector of $X$ corresponding to the feature set $S$, $S^c$ is the complement of the feature set $S$ and $\{X_S, X_{S^c}\} = X$ denotes the concatenation of $X_S$ and $X_{S^c}$.
\end{definition}
In other words, given a data point $\bfx$, the utility of the feature subset $S$ is defined as the impact on the model prediction, after ``removing'' the contribution from $S^c$ via integration with respect to the reference distribution $r$. These ``removal-based'' strategies are common in the explainability literature~\cite{covert2021explaining}. Nonetheless, the correct choice of the reference distribution has been a long-standing debate~\cite{chen2020true}. \citet{janzing2020feature} argued from a causality perspective that the feature marginal distribution should be used as the reference distribution, i.e. $r(X_{S^c}\mid X_S=x_S) = p(X_{S^c})$ where $p$ is the data distribution. On the other hand, ~\citet{frye2020shapley} disagreed by pointing out these ``marginal'' value functions ignore feature correlations and lead to unintelligible explanations in higher-dimensional data, and they instead advocate the use of conditional distribution as reference, i.e. $r(X_{S^c}\mid X_S=x_S) = p(X_{S^c}\mid X_S=x_S)$. Thus, there is no consensus and in fact, \citet{chen2020true} took a neutral stand and argued the choice depends on the application at hand. This also leads to design of value functions for specific problems, e.g. improving local estimation~\cite{ghalebikesabi2021locality}, incorporating causal knowledge~\cite{frye2019asymmetric,heskes2020causal} and modelling structured data~\cite{duval2021graphsvx}. In this paper, we will design an appropriate value function for preference learning and show that naive application of the existing value function to preference learning will lead to unintuitive results.

% We will later show in experiments that naive implementation of SHAP algorithm on preference function $g$ will lead to unintuitive results and 

% We will later in Sec.~\ref{sec: method} present our value function design for preference learning. 

%To the best of our knowledge, this is a novel contribution, as explainability on preference modelling has not yet been explored. 

\shortparagraph{Shapley value estimation.} Given a data point $\bfx$ and a model $f$, estimating Shapley values consist of two main steps: Firstly, for each feature subset $S \subseteq \Omega$, estimate the value function $\nu_{\bfx, f}(S)$ either by Monte Carlo sampling from the reference distributions $r$, or by utilising a model specific structure to speed up the estimation such as in \textsc{LinearSHAP}~\cite{vstrumbelj2014explaining}, \textsc{DeepSHAP}~\cite{lundberg2017unified}, \textsc{TreeSHAP}~\cite{lundberg2018consistent}, and \textsc{RKHS-SHAP}~\cite{chau2021rkhs}. The former sampling procedure is straightforward when $r$ is the marginal distribution, but computationally heavy and difficult when $r$ is the conditional distribution, as it involves estimating an exponential number of conditional densities~\cite{yeh2022threading}. Finally, after estimating the value functions, one can compute the Shapley values based on Eq.~\ref{eq: shapley} or by utilising the efficient weighted least square approach proposed by \citet{lundberg2017unified}. 

\shortparagraph{Estimating value functions when $f\in\cH_k$.} We give a review to the recently introduced \textsc{RKHS-SHAP} algorithm proposed by \citet{chau2021rkhs} as it is another core component for \textsc{Pref-SHAP}. \textsc{RKHS-SHAP} is a SV estimation method for functions in a given RKHS. It circumvents the need for any density estimation and utilises the arsenal of kernel mean embeddings~\cite{muandet2016kernel} to estimate the value functions non-parametrically. Assume $k$ takes a product kernel structure across dimensions, then for any $f \in \cH_k$, by applying the \textit{reproducing property}~\cite{paulsen2016introduction}, the value function can be decomposed as:
\begin{align}
    \nu_{\bfx, S}(f) &= \left\langle f,~\EE_{r\left(X_{S^c}\mid X_S=\bfx_S\right)}\left[k\left(\{X_S, X_{S^c}\}, \cdot \right)\mid X_S=\bfx_S\right] \right\rangle_{\cH_k} \\
    &= \left\langle f, k_{X_S} \otimes \mu_{r(X_\Sc \mid X_S=\bfx_S)} \right\rangle_{\cH_k},
\end{align}
where $k_{X_S}$ is the product of kernels belonging to the feature set $S$, and $\mu_{r(X_\Sc \mid X_S=\bfx_S)}:= \int k_{X_\Sc}r(X_\Sc\mid X_S=\bfx_S)dX_\Sc$ is the kernel mean embedding~\cite{muandet2016kernel} of the reference distribution $r$. Depending on the choice of the reference distribution, one recovers either the standard kernel mean embedding or the conditional mean embedding. This allows us to arrive at a closed form expression of the value function and circumvents the need for fitting an exponential number of conditional densities.

\vspace{-0.1cm}
\section{Proposed method: \textsc{Pref-SHAP}}
\vspace{-0.2cm}
\label{sec: method}

In this section, we will present \textsc{Pref-SHAP}, a new Shapley explainability toolbox designed to explain preferences by attributing contribution scores over item-level and context-level covariates for our preference models. Recall the likelihood model for C-GPM from Sec.~\ref{sec: preference_learning}: 
\begin{align}
    p\left(y \mid \bfu, \bfxleft, \bfxright\right) = \sigma\left(yg_U\left(\bfu, \bfxleft, \bfxright\right)\right),
\end{align}
where $g_U$ is the context-included preference function that denotes the strength of preference of item $\bfxleft$ over item $\bfxright$ under context $\bfu$. As there are two distinct sets of covariates present, we will propose two different value functions to capture the influences from items and context variables respectively, and show how they could be estimated non-parametrically using tools from the kernel methods literature, as in \textsc{RKHS-SHAP}. 

\vspace{-0.2cm}
\subsection{Preferential value function for items}
\vspace{-0.2cm}

To explain a general preference model $g:\cX\times\cX\to\RR$, we propose the following \textit{preferential value function for items}.
\begin{definition}[Preferential value function for items] Given a preference function $g\in\cH$, a pair of items $\left(\bfxleft, \bfxright\right) \in \cX\times \cX$ to compare, we define the \textit{preferential value function for items} as $\nu^{(p_I)}: \cX \times \cX\times[0,1]^d\times \cH \to \RR$ such that:
\begin{align}
    \nu^{(p_I)}_{\bfxleft, \bfxright, S}(g) = \EE_q\left[g(\{\Xleft_S,\Xleft_{S^c}\},\{\Xright_S,\Xright_{S^c}\})\mid \Xleft_S=\bfxleft_S, \Xright_S=\bfxright_S \right]
\end{align}
where expectation is taken over the reference $q\left(\Xleft_{S^c},\Xright_\Sc \mid \Xleft_S=\bfxleft_S, \Xright_S=\bfxright_S\right)$.
\vspace{-0.2cm}
\end{definition}

We note that $\nu^{(p_I)}$ is also applicable to the context-specific preference models. For example, applying $\nu^{(p_I)}$ to $g_\bfu := g_U(\bfu, \cdot, \cdot)$ allows one to quantify the item covariate's influences under a specific context $\bfu$, while applying $\nu^{(p_I)}$ to $\bar{g} := \EE_{p(U)}[g_U(U, \cdot, \cdot)]$ quantifies the average influence from each of the item covariates instead.

Similar to standard value functions, the influence of a feature set $S$ shared by the items $\bfxleft,\bfxright$ is measured as the impact on the preference model after ``removing'' contributions from features in $\Sc$, via integration with respect to some reference distribution $r$. Similar to $g$, this value function is skew-symmetric in its first two arguments, i.e. $\pnu(\bfxleft,\bfxright, S, g) = -\pnu(\bfxright,\bfxleft, S, g)$. This is justified, since features that ``encourage'' preference of $\bfxleft$ over $\bfxright$ should naturally be the ones that ``discourage'' preference of $\bfxright$ over $\bfxleft$ to ensure consistency. In this paper, we assume the items are i.i.d sampled from some distribution $p$, and we utilise the observational data distribution as reference as in \cite{frye2020shapley}, i.e. we take $r\left(\Xleft_{S^c},\Xright_\Sc \mid \Xleft_S=\bfxleft_S, \Xright_S=\bfxright_S\right)$ to be $p\left(\Xleft_{S^c}\mid \Xleft_S=\bfxleft_S\right)p\left(\Xright_{S^c}\mid \Xright_S=\bfxright_S\right)$. Although we decide here to use the observational distribution as the reference, the corresponding estimation procedure follows analogously if one instead uses the marginal distribution approach in \citet{janzing2020feature}.

\shortparagraph{Problems with direct application of SHAP to preference model $g$} A naive way of explaining with SHAP a general preference model $g$ which assumes no rankability would require concatenation of the items' covariates. Namely, we would set $\bfz = (\bfxleft,\bfxright)\in\RR^{2d}$ and then apply SHAP to the function $g(\bfz)$ directly, now giving $2d$ Shapley values for each observed preference, i.e. two Shapley values for each feature. Not only does this approach require us to consider a larger number of feature coalitions during computation (squaring the original amount), but it also ignores that $\bfxleft$ and $\bfxright$ in fact consist of the same features, leading to inconsistent explanations, i.e. that the same feature in $\bfxleft$ and $\bfxright$ has a different influence, hence giving different explanations simply due to the ordering of items. We illustrate these pitfalls of such a naive approach in Appendix \ref{appendix: naive}.

\shortparagraph{Empirical estimation of the preferential value function $\nu^{(p_I)}_{\bfxleft, \bfxright, S}(g)$} While the \textit{preferential value function} is general in the sense that it could be applied to any preference function $g$, we divert our attention to functions in $\cH_{k_E}$, where $k_E$ is the \textit{generalised preferential kernel} introduced in Sec~\ref{sec: preference_learning}. This allows us to adapt the recently introduced \textsc{RKHS-SHAP} to our settings, and we can thus circumvent learning an exponential number of conditional densities as in~\cite{frye2020shapley}. In the following segment, we prove the existence of the Riesz representation of the \textit{preferential value functional}, a necessary step to adapt the \textsc{RKHS-SHAP} framework to our setting.
\begin{proposition}[Preferential value functional for items]
Let $k$ be a product kernel on $\cX$, i.e. $k(\bfxleft,\bfxright) = \prod_{j=1}^d k^{(j)}(x^{(j)}, x'^{(j)})$. Assume $k^{(j)}$ are bounded for all $j$, then the Riesz representation of the functional $\nu^{(p)}_{\bfxleft, \bfxright, S}$ exists and takes the form:
\begin{align*}
    \nu^{(p)}_{\bfxleft, \bfxright, S} = \frac{1}{\sqrt{2}} \left(\cK(\bfxleft, S) \otimes \cK(\bfxright, S) - \cK(\bfxright, S) \otimes \cK(\bfxleft, S)\right)
\end{align*}
where $\cK(\bfx, S) = k_{S}(\cdot, \bfx_S)\otimes \mu_{X_{S^c}\mid X_S = \bfx_S}$ and $k_S(\cdot, \bfx_S) = \bigotimes_{j\in S}k^{(j)}(\cdot, x^{(j)})$ is the sub-product kernel defined analogously as $X_S$.

% \begin{align*}
% \frac{1}{\sqrt{2}}\Big[(k_{S}(\bfxleft_S)\otimes \mu_{X_{S^c}\mid \bfxleft_S})\otimes (k_S(\bfxright_S)\otimes \mu_{X_{S^c}\mid \bfxright_S}) - (k_S(\bfxright_S)\otimes \mu_{X_{S^c}\mid \bfxright_S}) \otimes (k_S(\bfxleft_S)\otimes \mu_{X_{S^c}\mid \bfxleft_S})\Big]
% \end{align*}
\end{proposition}
All proofs are included in the appendix. By representing the functionals as elements in the corresponding RKHS, we can now estimate the value function non-parametrically using kernel mean embeddings.
\begin{proposition}[Non-parametric Estimation]
\label{prop: pref-shap_esitmation}
Given $\hat{g} = \sum_{j=1}^m \alpha_j k_E((\bfxleft_j, \bfxright_j), \cdot)$, datasets $\bfXleft, \bfXright$, test items $\bfxleft, \bfxright$, the preferential value function at test items $\bfxleft, \bfxright$ for coalition $S$ and preference function $\hat{g}$ can be estimated as
\begin{align*}
    \hat{\nu}_{\small \bfxleft,\bfxright, S}^{(p_I)}(\hat{g}) = \balpha^\top\left(\Gamma(\bfXleft_S, \bfxleft_S)\odot \Gamma(\bfXright_S, \bfxright_S) - \Gamma(\bfXleft_S, \bfxright_S)\odot \Gamma(\bfXright_S, \bfxleft_S)\right),
\end{align*}
where $\Gamma(\bfXleft_S, \bfxleft_S) = \bfK_{\bfXleft_S, \bfxleft_S}\odot \bfK_{\bfXleft_{S^c}, \bfX_{S^c}}\bfK_{\bfX_S, \lambda}^{-1}\bfK_{\bfX_S, \bfxleft_S}, \bfK_{\bfX_S, \lambda} = \bfK_{\bfX_S, \bfX_S} + n\lambda I$, $\balpha = \{\alpha_j\}_{j=1}^m$ and $\lambda > 0$ is a regularisation parameter.
\end{proposition}
\vspace{-0.2cm}
\subsection{Preferential value function for contexts} 
\vspace{-0.2cm}

The influence an individual context feature in $U$ has on a C-GPM function $g_U$ can be measured by the following value function.

\begin{proposition}[Preferential value function for contexts] Given a preference function $g_U\in \cH_{k_E^U}$, denote $\Omega' = \{1,...,d'\}$, then the utility of context features $S'\subseteq \Omega'$ on $\{\bfu, \bfxleft,\bfxright\}$ is measured by $\nu^{(p_U)}_{\bfu, \bfxleft,\bfxright, S'}(g_U) = \EE[g_U\left(\{\bfu_S, U_{\Sc}\}, \bfxleft,\bfxright\right)\mid U_S=\bfu_S]$ where the expectation is taken over the observational distribution of $U$. Now, given a test triplet ($\bfu, \bfxleft, \bfxright)$, if $\hat{g}_U= \sum_{j=1}^m \alpha_j k_E^U\left((\bfu_j, \bfxleft_j, \bfxright_j), \cdot\right)$, the non-parametric estimator is:
\begin{align*}
    \hat{\nu}^{(p_U)}_{\bfu,\bfxleft,\bfxright, S'}(\hat{g}_U) &= \balpha^\top \left(\left( \bfK_{\bfU_{S'}, \bfu_{S'}} \odot \bfK_{\bfU_{S'^c}, \bfU_{S'^c}}\left(\bfK_{\bfU_{S'}, \bfU_{S'}} + m\lambda' I\right)^{-1}\bfK_{\bfU_{S'}, \bfu_{S'}}\right)\odot \Xi_{\bfxleft,\bfxright} \right)
\end{align*}
where $\Xi_{\bfxleft,\bfxright} = \left(\bfK_{\bfXleft, \bfxleft}\odot \bfK_{\bfXright, \bfxright} - \bfK_{\bfXright, \bfxleft}\odot \bfK_{\bfXleft, \bfxright}\right)$.
\end{proposition}

\begin{table}[t!]
\center
\caption{A summary of how our preference value functions can tackle different explanation tasks}
\resizebox{\columnwidth}{!}{
\begin{tabular}{|c|c|c|c|}
\hline
Candidate & Explanation of interest & Value function & Preference function\\ \hline
    $\bfxleft, \bfxright$      &    Which item features contributed most to this duel?                     &       $\pnu_{\bfxleft,\bfxright, S}$ &         $g$, $\EE_{U}[g_U(U, \cdot, \cdot)]$\\  \hline
    $\bfxleft$      &       Which item features contributed most to $\bfxleft$'s matches?                  &    $\frac{1}{n}\sum_{i=1}^n \pnu_{\bfxleft,\bfx_i, S}$            & $g$, $\EE_{U}[g_U(U, \cdot, \cdot)]$ \\ \hline
    $\bfu, \bfxleft,\bfxright$      &   Which context features contributed most to this duel?                     &     $\nu^{(p_U)}_{\bfu, \bfxleft,\bfxright, S}$           & $g_U$ \\\hline
     $\bfu$& Which context features contributed most on average? &$\frac{1}{m}\sum_{j=1}^m \nu^{(p_U)}_{\bfu,\bfxleft_j,\bfxright_j, S'}$ & $g_U$\\ \hline
\end{tabular}
}
\label{table: interrogation}
\vspace{-0.4cm}
\end{table}

Analogously, the average influence of a specific context feature can be computed by taking an average over all pairs of matches, i.e. by using a modified value function $\frac{1}{m}\sum_{j=1}^m \nu^{(p_U)}_{\bfu,\bfxleft_j,\bfxright_j, S'}(\hat{g}_U)$. We summarise different ways to modify the proposed preferential value functions to interrogate the preference models in Table~\ref{table: interrogation}. 

\shortparagraph{Computational complexity of \textsc{Pref-SHAP}} When computing GPM, it is fundamentally a \emph{kernel ridge regression} (KRR), which naïvely has complexity $\mathcal{O}(n^3)$. There exists a multitude of approximation techniques for KRR, the most common type being the Nyström approximation \cite{JMLR:v6:drineas05a}. For all our experiments, we use FALKON~\cite{Meanti2020KernelMT}, a large-scale library for solving kernel logistic regression using preconditioned conjugate gradient descent and Nystr\"om approximations. FALKON has a computational complexity of $\mathcal{O}(n\sqrt{n})$, which effectively becomes the complexity for GPM when using FALKON. As the value function for GPM requires estimating conditional mean embeddings, which in turn also are KRR's, one can appeal to FALKON again to reduce complexity to $\mathcal{O}(n\sqrt{n})$. We summarize the procedure of \textsc{Pref-SHAP} in Algorithm \ref{main_alg}.
\begin{algorithm}[htb!]
\caption{\textsc{Pref-SHAP}}
\label{main_alg}
\begin{algorithmic}[1]
\Require Solution $\boldsymbol{\alpha}$, datasets $\bfXleft,\bfXright,\bfX, \bfU$, test items $\bfxleft,\bfxright,\mathbf{u}$, batch size $n_b$, number of coalition samples $n_S$, context-specific flag $\texttt{cflg}$
\State Compute effective dimension $d_{\text{eff}}$ := Number of features with variance greater than 0. 
\State Compute coalitions $\mathcal{S}=\{S_1,\hdots, S_{n_S}\}$, form binary matrix $\mathbf{Z} \in \{0,1\}^{n_S,d_{\text{eff}}}$ from $\mathcal{S}$, and compute weights $\mathbf{W}=[w_1,\hdots,w_{n_S}]$ with $w_i=\frac{d-1}{ {{d}\choose{|S_i|}} |S_i|(d-|S_i|)}$.
\For { batch $\mathcal{S}_b$ in $\mathcal{S}$}
\State \textbf{if} \texttt{cflg} Take $S,S^c$ of $\bfXleft,\bfXright,\bfX,\bfxleft,\bfxright$ \textbf{else} Take $S',S'^c$ of $\bfU,\bfu$ 
\State \textbf{if} \texttt{cflg} Compute $\bfK_{\bfX_S, \lambda}^{-1}[\bfK_{\bfX_S, \bfxleft_S},\bfK_{\bfX_S, \bfxright_S }]$ \textbf{else} $\left(\bfK_{\bfU_{S'}, \bfU_{S'}} + m\lambda' I\right)^{-1}\bfK_{\bfU_{S'}, \bfu_{S'}}$ using \texttt{BatchedCGD}
\State \textbf{if} \texttt{cflg} Compute $ \hat{\nu}_{\small \bfxleft,\bfxright, \mathcal{S}_b}^{(p)}(\hat{g})$ \textbf{else} $  \hat{\nu}_{\bfu,\bfxleft,\bfxright, \mathcal{S}_b'}(\hat{g}_U)$
\EndFor
\State \textbf{if} \texttt{cflg} Set $\mathbf{v}_x=\{ \hat{\nu}_{\small \bfxleft,\bfxright, \mathcal{S}_b}^{(p)}(\hat{g})\}^{B}_{b=1}$ \textbf{else} $\mathbf{v}_x=\{\hat{\nu}_{\bfu,\bfxleft,\bfxright, \mathcal{S}_b'}(\hat{g}_U)\}^{B}_{b=1}$
\State Calculate Shapley values $\boldsymbol{\beta}_{x}=\left(\mathbf{Z}^{\top}\mathbf{W} \mathbf{Z}\right)^{-1} \mathbf{Z}^{\top} \mathbf{W} \mathbf{v}_{x}$\\
\Return $\boldsymbol{\beta}_{x}$
\end{algorithmic}
\end{algorithm}
We further detail computational details pertaining to computing coalitions $S$ and batched conjugate gradient descent (\texttt{BatchedCGD}) in Appendix \ref{compute_stuff}.

\vspace{-0.2cm}
\section{Experiments}
\vspace{-0.2cm}

%1. No shapley values on preferences, actually it's just Shapley on utility functions. Utility: algebraically the difference of utility; this is problematic, particularly when explaining since it's too simple. Explaining these utility differences often not very fruitful.
%2. Since we care about preference, we want to explain the bi input covariates together with contests covariates. 

The main focus of our experiments is to illustrate the difference between explaining GPM (\textsc{Pref-SHAP}) and applying SHAP to UPM, thus highlighting the difference in explaining the mechanism of eliciting preferences and explaining the utility. %When we explain UPM, as 
When we explain UPM, we first explain how items $\bfxleft,\bfxright$ affect their utilities $f(\bfxleft),f(\bfxright)$. Explaining the utility corresponds to calculating the value functions of the utilities $\nu_{\bfxleft, S}(f)$ and $\nu_{\bfxright, S}(f)$. By linearity of SHAP values~\cite{lundberg2017unified} and the simple structure relating preference and utilities in UPM, we can explain UPM by subtracting the Shapley values of $\bfxleft$ with $\bfxright$. However, this type of explanation is only correct when data is rankable, which seldom happens in practice, thus motivating \textsc{Pref-SHAP}.

% as $\nu_{\bfxleft, S}(f)-\nu_{\bfxright, S}(f)$.

We apply \textsc{Pref-SHAP} to unrankable synthetic and real-world datasets to connect theory with practice. We split data, i.e. matches with their outcomes, into train (80\%), validation (10\%), and test (10\%) and explain the model on a random subset of the data. The hyperparameters for the kernels are selected using gradient descent, based on the proposed method in ~\cite{https://doi.org/10.48550/arxiv.2201.06314}. We first generate synthetic duelling data where performance can be compared against ground truth, to demonstrate that \textsc{Pref-SHAP} is capable of identifying the relevant features.

% For a better comparison, we normalize all Shapley values with the  
% \begin{wrapfigure}[24]{r}{0.4\textwidth}
%   \begin{center}
%     \includegraphics[width=0.4\textwidth]{teaserplot_gpgp/bee_plot_lamb=0.0_lasso.png}
%         \includegraphics[width=0.4\textwidth]{teaserplot_pgp/bee_plot_lamb=0.0_lasso.png}
%   \end{center}
%     \vspace{-10pt}
%   \caption{\textsc{Pref-SHAP} (top) correctly finds that fire Pokémon are weak to water. Explaining utility based preference models (bottom) incorrect suggest that fire Pokémon can have a slight advantage against when battling water Pokémon!}
% \end{wrapfigure}

\begin{wrapfigure}[16]{r}{0.2\textwidth}
\vspace{-0.7cm}
  \begin{center}
    \includegraphics[width=0.2\textwidth]{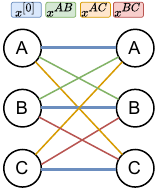}
  \end{center}
  \caption{An illustration of our simulation: each edge corresponds to the variable that dictates the comparison based on the colour.}
  \label{abc}
\end{wrapfigure}

\shortparagraph{Synthetic data} We first consider a synthetic experiment with unrankable duelling data. We generate the items by first sampling $1000$ item covariates $[x^{[0]}_i,x^{AB}_i,x^{AC}_i,x^{BC}_i]=:\mathbf{p}_i \in \mathbb{R}^4 \sim \mathcal{N}(0,\mathbf{I}_4)$. We associate each item with a cluster membership $c_i \in \{A,B,C\}$, where the assignment is randomly chosen for each item with equal probability. 
We then form the full item covariate by concatenating $\mathbf{p}_i$ with one-hot encoded $c_i$ as $\bfx_i = [\mathbf{p}_i, \texttt{one\_hot}(c_i)]$. $40000$ matches between randomly chosen pairs of items are conducted by the following mechanism: match outcomes are decided based on the underlying cluster membership of the items. For example, if an item from cluster $A$ competes against an item from cluster $B$, the winner is decided by their inter-cluster covariate $x^{AB}$, i.e. $i \preceq j$ if $x_j^{AB} \geq x_i^{AB}$. When the match is between members of the same cluster, it is dictated by the maximum among the within-cluster variable, i.e. $\max(x^{[0]}_i,x^{[0]}_j)$. See Fig.~\ref{abc} for an illustration. As no clusters have any advantage over the others, the data is not rankable, and we expect the inter-cluster covariates $x^{AB}, x^{AC}, x^{BC}$ to have similar explanations on average, but significantly different from each other when we examine local explanations.

\begin{figure}[htb!] %All plots between 1-5
\centering
    \begin{subfigure}[H]{0.25\linewidth}
    \centering
    \rotatebox{0}{\scalebox{0.75}{\textsc{Pref-SHAP}}}
    \end{subfigure}%
    \begin{subfigure}[H]{0.25\linewidth}
    \centering
    \rotatebox{0}{\scalebox{0.75}{SHAP for UPM}}
    \end{subfigure}%
   \begin{subfigure}[H]{0.25\linewidth}
    \centering
    \rotatebox{0}{\scalebox{0.75}{\textsc{Pref-SHAP}}}
    \end{subfigure}%
    \begin{subfigure}[H]{0.25\linewidth}
    \centering
    \rotatebox{0}{\scalebox{0.75}{SHAP for UPM}}
    \end{subfigure}
    \begin{subfigure}{0.495\textwidth}
        \centering
        \includegraphics[width=0.5\linewidth]{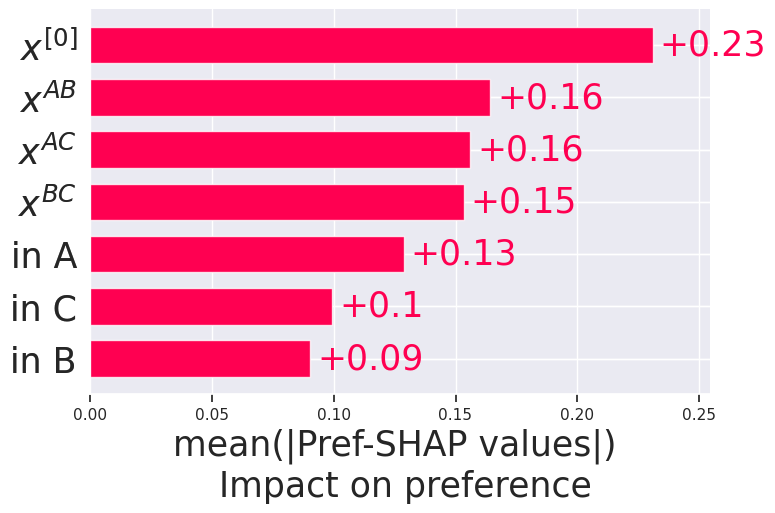}%
                \includegraphics[width=0.5\linewidth]{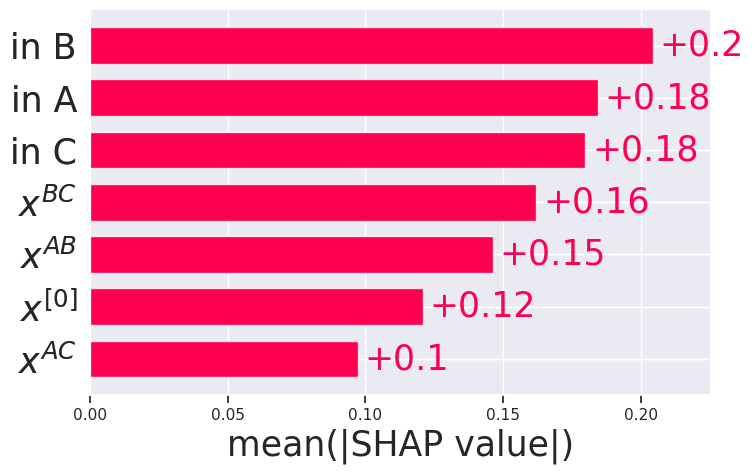}
        % \includegraphics[width=\linewidth]{test/bee_plot_lamb=0.0_ridge.png}
        % \caption{Barplots for the Pokémon dataset. Shapley values calculated on test data. \\ \quad}
        % \label{synt_barplot}
    \end{subfigure}%
    \hfill
    \begin{subfigure}{0.495\textwidth}
        \centering
        \includegraphics[width=0.5\linewidth]{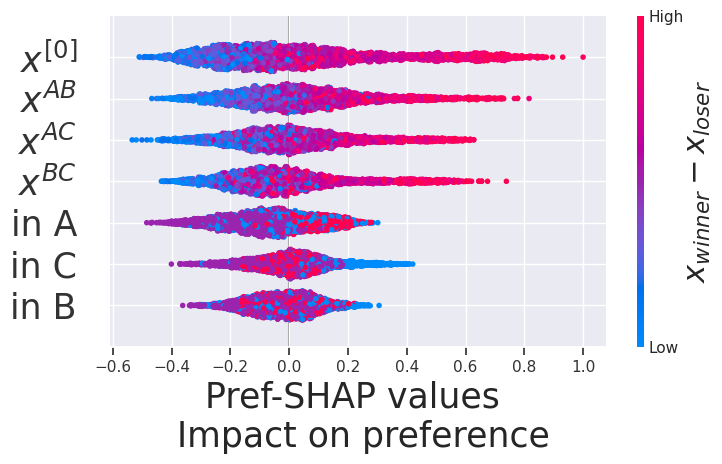}%
                \includegraphics[width=0.5\linewidth]{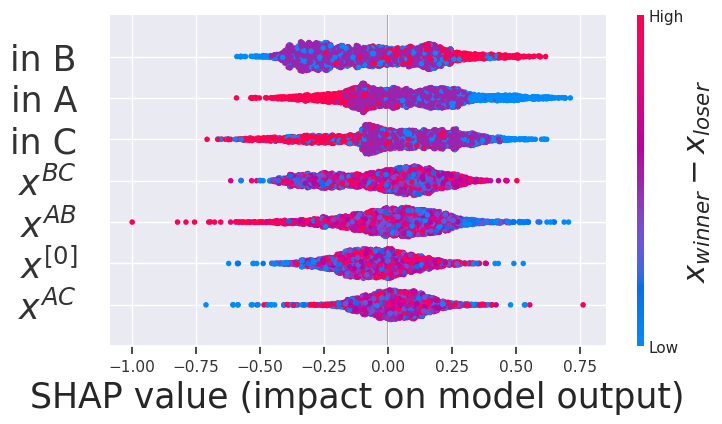}
        % \caption{Beehive plots for the Pokémon dataset. \textsc{Pref-SHAP} captures that both speed and type matter, while explaining rank only captures the type importance.}
        % \label{synt_beeplot}
    \end{subfigure}
     \caption{\small{Bar and Beehive plots for global explanations on the synthetic dataset.}}
  \label{synt_beeplot_global}
  \vspace{-0.3cm}
\end{figure}

\begin{figure}[htb!] %All plots between 1-5

\centering
    \begin{subfigure}[H]{0.25\linewidth}
    \centering
    \rotatebox{0}{\scalebox{0.75}{\textsc{Pref-SHAP}}}
    \end{subfigure}%
    \begin{subfigure}[H]{0.25\linewidth}
    \centering
    \rotatebox{0}{\scalebox{0.75}{SHAP for UPM}}
    \end{subfigure}%
   \begin{subfigure}[H]{0.25\linewidth}
    \centering
    \rotatebox{0}{\scalebox{0.75}{\textsc{Pref-SHAP}}}
    \end{subfigure}%
    \begin{subfigure}[H]{0.25\linewidth}
    \centering
    \rotatebox{0}{\scalebox{0.75}{SHAP for UPM}}
    \end{subfigure}
    \begin{subfigure}{0.495\textwidth}
        \centering
        \includegraphics[width=0.5\linewidth]{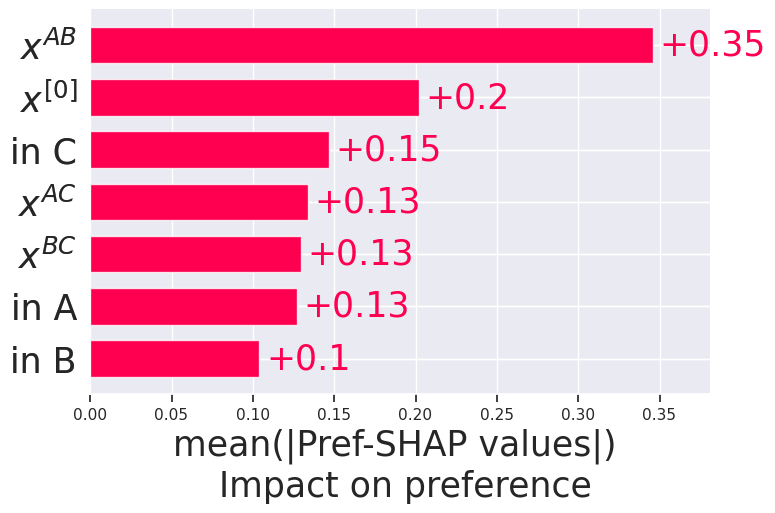}%
                \includegraphics[width=0.5\linewidth]{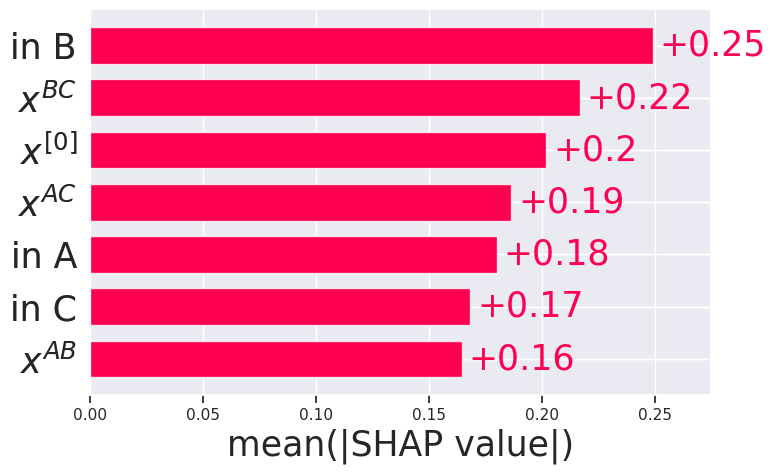}
        % \includegraphics[width=\linewidth]{test/bee_plot_lamb=0.0_ridge.png}
        % \caption{Barplots for the Pokémon dataset. Shapley values calculated on test data. \\ \quad}
        % \label{synt_barplot}
    \end{subfigure}%
    \hfill
    \begin{subfigure}{0.495\textwidth}
        \centering
        \includegraphics[width=0.5\linewidth]{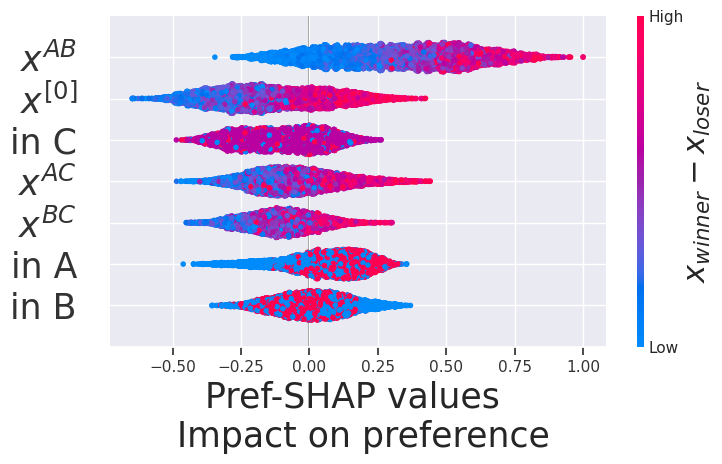}%
                \includegraphics[width=0.5\linewidth]{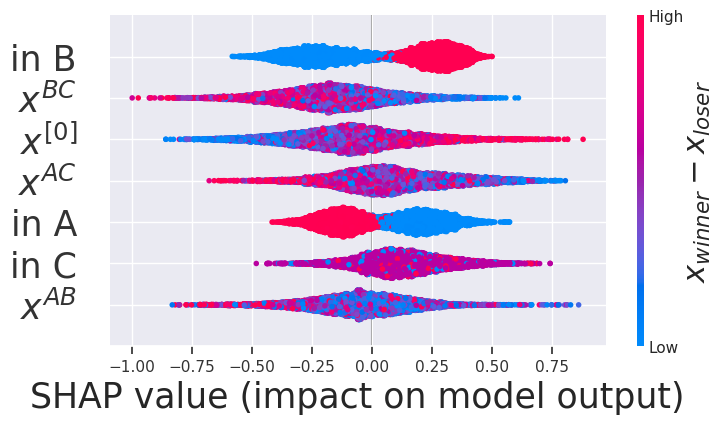}
        % \caption{Beehive plots for the Pokémon dataset. \textsc{Pref-SHAP} captures that both speed and type matter, while explaining rank only captures the type importance.}
        % \label{synt_beeplot}
    \end{subfigure}
     \caption{\small{Bar and Beehive plots for grouped local explanations on the synthetic dataset (Cluster $A$ vs $B$).}}
  \label{synt_beeplot_local}
 \vspace{-0.3cm}
\end{figure}

% We then randomly sample 40000 matches between pairs $i,j$. The matches are decided by comparing features based on cluster membership, according to \Cref{abc}. As an example, if members $i,j$ from $A,B$ respectively competed, the winner would be decided by the maximum between $x^{AB}_i$ and $x^{AB}_j$. If the match is between members of the same cluster, then the winner is the maximum between $x^{[0]}_i$,$x^{[0]}_j$. As no player will have an obvious advantage over other items, the dataset is unrankable. The dataset also has local structures following the clustered setup. 
We consider both global and \emph{grouped-local} explanations of the synthetic dataset in \Cref{synt_beeplot_global} and \Cref{synt_beeplot_local} respectively. In the global explanations, we explain all matches regardless of the cluster membership, while in the grouped-local explanations we only explain matches between items from $A$ against items from $B$. For more grouped-local explanations on different cluster pairs, we refer to appendix \ref{local_exp_appendix}.

% The color bar of the beehive plots indicate the magnitude of the feature, where red values indicate higher values and blue indicate lower values.  $\bfx_{\text{winner}}-\bfx_{\text{loser}}$, as this retains the interpretation of linear magnitude, similar to the single-input case. 

\shortparagraph{Interpreting the simulation explanations.} The beehive plots showcase the recovered \textsc{Pref-SHAP} values, where the bar plots demonstrated the average \textsc{Pref-SHAP} values for each feature. The colour in the beehive plots indicates the magnitude of the difference between the corresponding features of the winner and of the loser in that match. For example, a red point in a beehive plot for feature $d$ indicates that the difference $x_{winner}^{(d)} - x_{loser}^{(d)}$ is large.

Fig.~\ref{synt_beeplot_global} illustrates the explanation results for the global synthetic experiments. We see that \textsc{Pref-SHAP} identified the within-cluster variable $x^{[0]}$ as the most important, which is a consequence of the fact that the largest number of matches are played between the items of the same cluster (cf. Fig. \ref{abc} where there are three blue lines and two lines of each of the other colours). The three inter-cluster variables contributed similarly according to \textsc{Pref-SHAP}, which, by symmetry, should be the case. Furthermore, the correct battle mechanism is captured by \textsc{Pref-SHAP} but not UPM, as we see that the large \textsc{Pref-SHAP} values for each feature are red in the beehive plot. This indicates that items with larger value are more likely to win against items with lower value in the corresponding features. %This aligns with the experiment design as well.
In contrast, SHAP for UPM does not recover this insight.

The explanations for the matches between items from $A$ against items from $B$, are shown in Fig~\ref{synt_beeplot_local}. Here, $x^{AB}$ is correctly picked as the relevant feature in these matches with \textsc{Pref-SHAP}, but not with SHAP for UPM. We see again that there is a clear tendency that large \textsc{Pref-SHAP} values are red for feature $x^{AB}$, showing that \textsc{Pref-SHAP} once again captures the designed gaming mechanism -- which is not the case in SHAP for UPM. Intuitively, even though SHAP for UPM allows local explanations, it does so based on a \emph{global utility}, which fails completely in a non-rankable case. 

% \textsc{Pref-SHAP} correctly recovers that high differences between the winner and loser in the first 4 features are the most important. SHAP for UPM fails to accurately recover this insight as it doesn't find a pattern where high differences between winner and loser have positive outcome on model.

% \shortparagraph{Interpreting the explanations.} The color bar of the beehive plots indicate the signed difference in values of the corresponding features between $\bfxleft$ and $\bfxright$, i.e. the colour intensity of the Shapley values of feature $x^{[0]}$ is proportional to $x_i^{[0]} - x_j^{[0]}$ when player $i, j$ are competing. Another example is that, in \Cref{synt_beeplot_local}, 

%  \textsc{Pref-SHAP} correctly recovers that the second feature $x^{AB}$ is the most important, correctly relating high differences to positive correlation with winning. SHAP for UPM is unable to correctly identify the important feature.

\shortparagraph{Real-world explanations} For our real-world datasets, we consider publicly available datasets \textit{Chameleon}, \textit{Pokémon} and \textit{Tennis}. We provide descriptive statistics of these datasets in \Cref{dat_summary} and give their brief descriptions below. Appendix \ref{website_stuff} contains further large scale experiments on an additional dataset consisting of user-item interactions on a fashion retail website.

\begin{table}[t!]
    \centering
        \caption{Dataset summary}
    \label{dat_summary}
    \resizebox{\linewidth}{!}{%

\begin{tabular}{cccccccc} 
\hline
          Dataset & $N_{\text{Matches}}$ & $N_{\text{items}}$ & $N_{\text{Context}}$ & $D_{\text{continuous}}$ & $D_{\text{binary}}$ & $D^{\text{Context}}_{\text{continuous}}$ & $D^{\text{Context}}_{\text{binary}}$ \\ \hline
\textit{Synthetic} & $40000$                & $1000$                 & -                   & $4$                       & $3$                   & -                                      & -                                  \\
\textit{Chameleon} & $106$                  & $35$                   & -                   & $7$                       & $19$                  & -                                      & -                                  \\
\textit{Pokémon}   & $60000$                & $800$                  & -                   & $7$                       & $0$                   & -                                      & -                                  \\
\textit{Tennis}    & $95359$                & $3483$                 & $4114$~(tournaments)    & $4$                       & $7$                  & $0$                                        & $6$                                    \\
\hline
\end{tabular}
}
\vspace{-0.5cm}
\end{table}

% \textit{Website}   & $85144$                & $20626$                & $129117$~(users)        & $0$                       & $93$                  & $1$                                        & $4$                                    \\

\textit{The Chameleon} dataset~\cite{chameleon_cite} considers 106 contests between 35 male dwarf chameleons. Physical traits of the chameleons are measured such as the \textit{height of their casque}, \textit{length of their jaw}, \textit{body mass} etc. According to~\cite{chameleon_cite}, they fitted a linear Bradley Terry model and examined the coefficients to deduce that \textit{casque height} (ch.res) and \textit{relative area of the flank patch} (prop.patch) positively affected the fighting ability the most. \textit{The Pokémon} dataset considers 60000 Pokémon battles among 800 Pokémon. Pokémon have different characteristics such as \textit{attack power}, \textit{speed}, \textit{health} etc. The Pokémon further has at least one different \textit{type} such as \textit{Electric, Water, Fire}, etc. Certain types have advantages and disadvantages against each other, for instance, fire Pokémon are weak to water-based attacks (receiving twice the damage) and as a result have a disadvantage against water Pokémon. 

\textit{The Tennis} dataset considers professional tennis matches between 1991 and 2017 in all major tournaments each year. The data is provided publicly by ATP World Tour~\cite{tennis_data}. Features such as \textit{birthyear}, \textit{weight}, \textit{height} etc are included about each tennis player together with context details of the match such as the court being indoor or outdoor and what surface the match is being played on.

The above datasets are not rankable, and we validate this claim by comparing GPM performance against UPM in \Cref{GPL_vs_UPL}, together with the estimated rankability measure \emph{SpecR} proposed in~\cite{anderson2019rankability} for each dataset. \emph{SpecR} measures the similarity of the data to a complete dominance graph (i.e. rankable data). It takes values between 0 and 1 with values close to 1 being evidence in support of rankability. For the Tennis data where there are additional relationships with the context~(tournaments), we estimate the average \emph{SpecR} of each tournament.
\begin{table}[t!]
\centering
 \caption{GPM vs UPM. Mean and standard deviations of performance averaged over 5 runs.}
 \label{GPL_vs_UPL}
 \resizebox{\linewidth}{!}{%

\centering
\begin{tabular}{l|cc|cc|cc|cc|} 
\hline
         & \multicolumn{2}{c|}{Synthetic}                                             & \multicolumn{2}{c|}{Chameleon}                                             & \multicolumn{2}{c|}{Pokémon}                                               & \multicolumn{2}{c|}{Tennis}                                                 \\
         & GPM                                  & UPM                                 & GPM                                  & UPM                                 & GPM                                  & UPM                                 & C-GPM                                & UPM                                  \\ 
\hline
Test AUC & \multicolumn{1}{l}{$0.98_{\pm0.00}$} & \multicolumn{1}{l|}{$0.71_\pm0.01$} & \multicolumn{1}{l}{$0.92_{\pm0.07}$} & \multicolumn{1}{l|}{$0.80_{\pm0.07}$} & \multicolumn{1}{l}{$0.86_{\pm0.00}$} & \multicolumn{1}{l|}{$0.82_{\pm0.00}$} & \multicolumn{1}{l}{$0.58_{\pm0.02}$} & \multicolumn{1}{l|}{$0.52_{\pm0.02}$}  \\
SpecR    & \multicolumn{2}{c|}{$0.09$}                                                & \multicolumn{2}{c|}{$0.24$}                                                & \multicolumn{2}{c|}{$0.20$}                                                & \multicolumn{2}{c|}{$0.13_{\pm 0.07}$}                                      \\
\hline
\end{tabular}
}
\vspace{-0.5cm}
\end{table}
Both the superior performance of GPM over UPM and the low \emph{SpecR} measures suggest that the datasets are generally not rankable, which points to limitations of explaining preferences via utility-based modeling.

\begin{figure}[htb!] %All plots between 1-5
\centering
    \begin{subfigure}[H]{0.25\linewidth}
    \centering
    \rotatebox{0}{\scalebox{0.75}{\textsc{Pref-SHAP}}}
    \end{subfigure}%
    \begin{subfigure}[H]{0.25\linewidth}
    \centering
    \rotatebox{0}{\scalebox{0.75}{SHAP for UPM}}
    \end{subfigure}%
   \begin{subfigure}[H]{0.25\linewidth}
    \centering
    \rotatebox{0}{\scalebox{0.75}{\textsc{Pref-SHAP}}}
    \end{subfigure}%
    \begin{subfigure}[H]{0.25\linewidth}
    \centering
    \rotatebox{0}{\scalebox{0.75}{SHAP for UPM}}
    \end{subfigure}
    \begin{subfigure}{0.495\textwidth}
        \centering
        \includegraphics[width=0.5\linewidth]{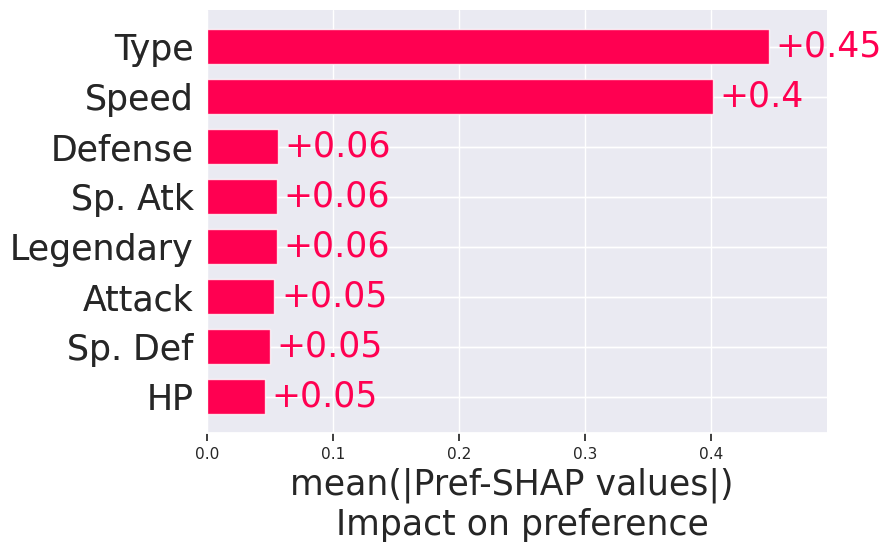}%
                \includegraphics[width=0.5\linewidth]{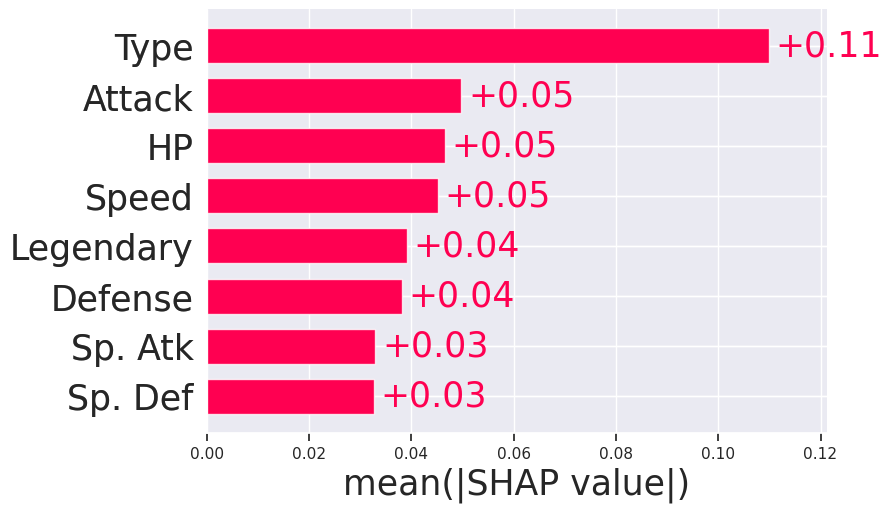}
        % \includegraphics[width=\linewidth]{test/bee_plot_lamb=0.0_ridge.png}
        % \caption{Barplots for the Pokémon dataset. Shapley values calculated on test data. \\ \quad}
        % \label{synt_barplot}
    \end{subfigure}%
    \hfill
    \begin{subfigure}{0.495\textwidth}
        \centering
        \includegraphics[width=0.5\linewidth]{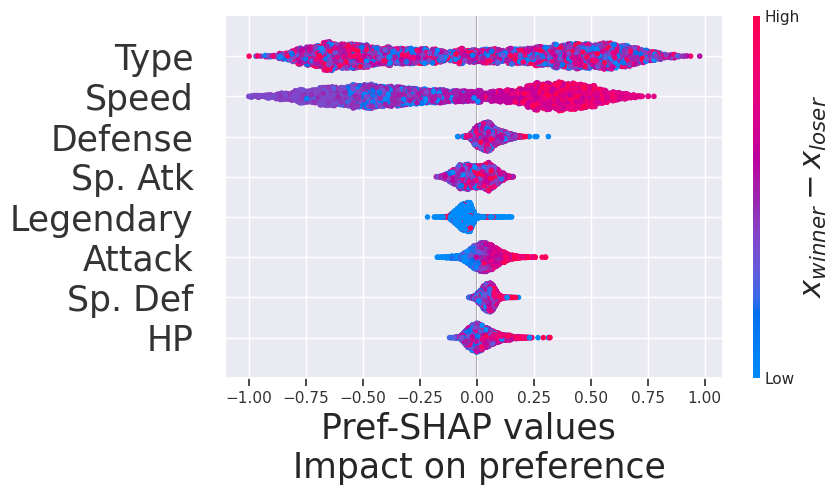}%
                \includegraphics[width=0.5\linewidth]{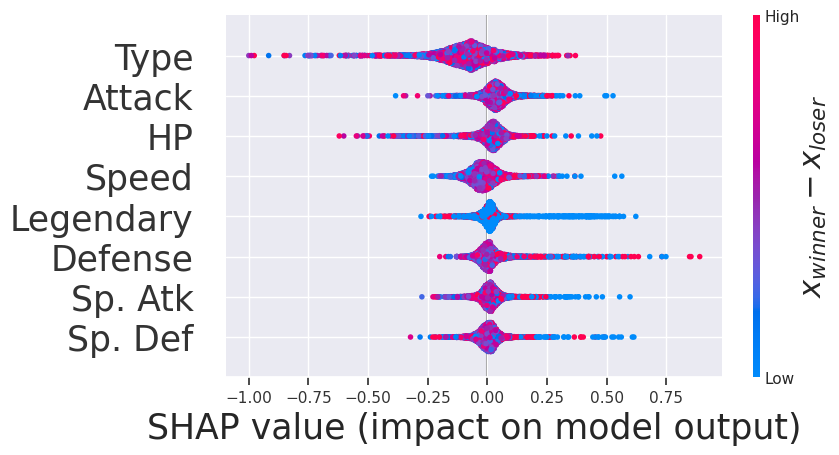}
        % \caption{Beehive plots for the Pokémon dataset. \textsc{Pref-SHAP} captures that both speed and type matter, while explaining rank only captures the type importance.}
        % \label{synt_beeplot}
    \end{subfigure}
     \caption{\small{Bar and Beehive plots for the Pokémon dataset. \textsc{Pref-SHAP} captures that both speed and type matter, while SHAP for UPM only captures the type importance.}}
  \label{poke_plots}
  \vspace{-0.3cm}
\end{figure}

\shortparagraph{Explaining Pokémon battles.} We first consider standard dueling data for explaining preferences. We explain the learned preferences and learned differences in utilities on the Pokémon dataset in \Cref{poke_plots}. In this dataset, we have summed the Shapley values for \emph{Type} features.

\begin{wrapfigure}[20]{r}{0.3\textwidth}
\vspace{-0.3cm}
  \begin{center}
    \includegraphics[width=0.3\textwidth]{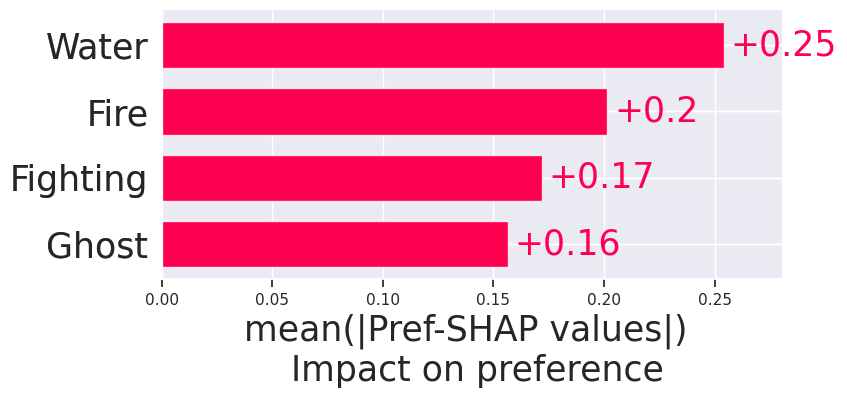}
        \includegraphics[width=0.3\textwidth]{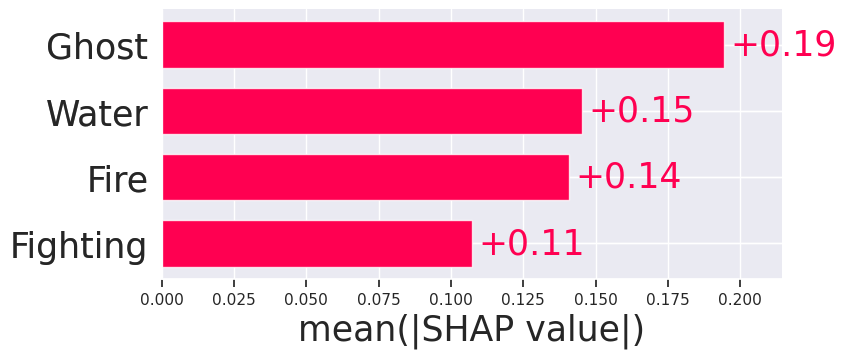}
  \end{center}
  \vspace{-0.2cm}
  \caption{\small Explaining matches between 4 types of Pokémon, among them only fire and water has a type disadvantage/advantage against each other. \textsc{Pref-SHAP} (top) correctly identifies that fire and water are the most important, while water and fire are not deemed most important by SHAP for UPM.}
  \label{fig: type_adv}
\end{wrapfigure}

We see that explaining general preferences provides further insight than just explaining the difference in utility functions. In particular, SHAP for UPM does not capture the additional importance of \emph{Speed} in winning battles. As higher (more red) values of differences in speed $x^{\text{speed}}_{\text{winner}}-x^{\text{speed}}_{\text{loser}}$ have positive impact on the outcome, we conclude that having higher speed than your opponent is advantageous besides a type advantage. This insight is aligned with the ``Sweeper'' strategy~\cite{sweeper_strat}, where one would employ a leading Pokémon with very high speed and attack to attempt downing the opponent before they can strike back. In Fig.~\ref{fig: type_adv}, we see \textsc{Pref-SHAP} can also capture the correct type advantage/disadvantages among the Pokémon, but not SHAP for UPM.

%  For the Chameleon dataset, both \textsc{RKHS-SHAP} and \textsc{SHAP} on UPM deemed \textit{prop.patch} most important. 

\shortparagraph{Explaining Chameleon contests.} We find that UPM's explanations are more aligned with~\cite{chameleon_cite}'s findings~(\textit{prop.path} and \textit{ch.res} are the most important features), which is unsurprising since the Bradley Terry model used in~\cite{chameleon_cite} is also a utility based model. However, since GPM gives a much better predictive performance than UPM (Test AUC $0.92$ v.s. $0.80$), we believe \textsc{Pref-SHAP}'s explanations are also insightful. In fact, \textsc{Pref-SHAP} discovers that having larger \textit{jaw sizes} (jl.res) than your opponent have a significant negative effect on match outcome, a previously undiscovered mechanism from~\cite{chameleon_cite}. We verify this finding in Appendix \ref{cham_proof} by applying \textsc{Pref-SHAP} to GPM trained on multiple folds of the Chameleon dataset and consistently find that high values of the \textit{jaw size} (jl.res) variable have a negative impact on the outcome. 

% This further motivates the choice of GPM as the default choice for both modelling and explaining preferences.

% \shortparagraph{Explaining Chameleon contests.} For the Chameleon dataset, both \textsc{Pref-SHAP} and SHAP on the UPM model are able to recover the same insights from~\cite{chameleon_cite}, where they highlight the importance of \textit{relative area of the flank patch} (prop.patch) by examining the linear coefficient from a Bradley-Terry model. Additionally, \textsc{Pref-SHAP} discovers that having larger \textit{jaw sizes} (jl.res) than your opponent have a significant negative effect on match outcome. We show in Appendix \ref{cham_proof} that when we apply \textsc{Pref-SHAP} to GPM trained on different folds of the Chameleon dataset, we consistently find that high values of the \textit{jaw size} (jl.res) variable have a negative impact on the outcome. This further motivates the choice of GPM as the default choice for both modelling and explaining preferences.

%  Comparing \textsc{Pref-SHAP} to explaining rank on the Chameleon dataset in \Cref{chameleon_plots} we find that explaining rank recovers the same insights from~\cite{chameleon_cite}, highlighting the importance of \textit{casque} (ch.res) and \textit{relative area of the flank patch} (prop.patch), when they applied a Bradley-Terry model. Interestingly, \textsc{Pref-SHAP} in additional also finds a strong negative signal for larger jaw sizes (jl.res), something previously unknown. 

\begin{figure}[htb!] %All plots between 1-5

\centering
    \begin{subfigure}[H]{0.25\linewidth}
    \centering
    \rotatebox{0}{\scalebox{0.75}{\textsc{Pref-SHAP}}}
    \end{subfigure}%
    \begin{subfigure}[H]{0.25\linewidth}
    \centering
    \rotatebox{0}{\scalebox{0.75}{SHAP for UPM}}
    \end{subfigure}%
   \begin{subfigure}[H]{0.25\linewidth}
    \centering
    \rotatebox{0}{\scalebox{0.75}{\textsc{Pref-SHAP}}}
    \end{subfigure}%
    \begin{subfigure}[H]{0.25\linewidth}
    \centering
    \rotatebox{0}{\scalebox{0.75}{SHAP for UPM}}
    \end{subfigure}
    \begin{subfigure}{0.495\textwidth}
        \centering
        \includegraphics[width=0.5\linewidth]{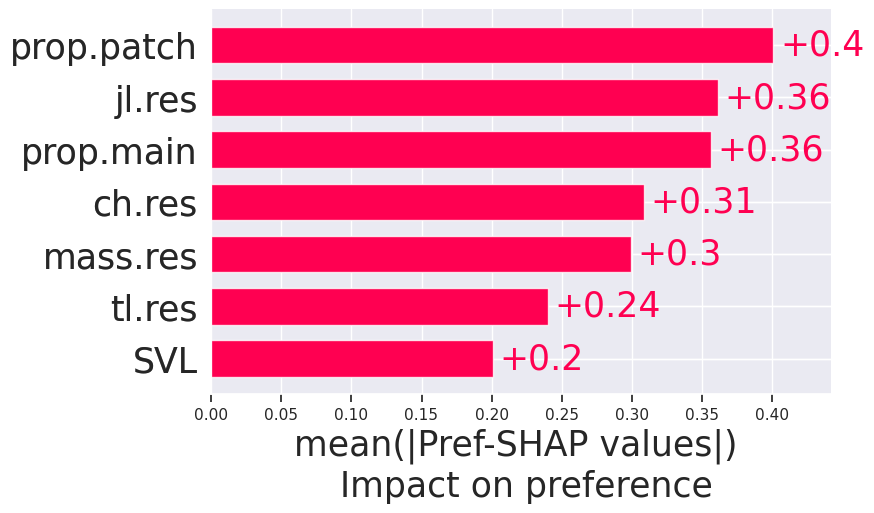}%
                \includegraphics[width=0.5\linewidth]{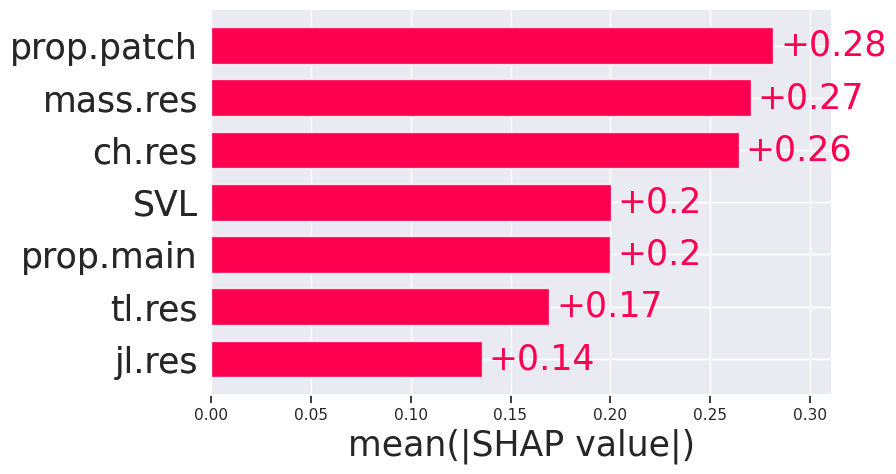}

    \end{subfigure}%
    \hfill
    \begin{subfigure}{0.495\textwidth}
        \centering
        \includegraphics[width=0.5\linewidth]{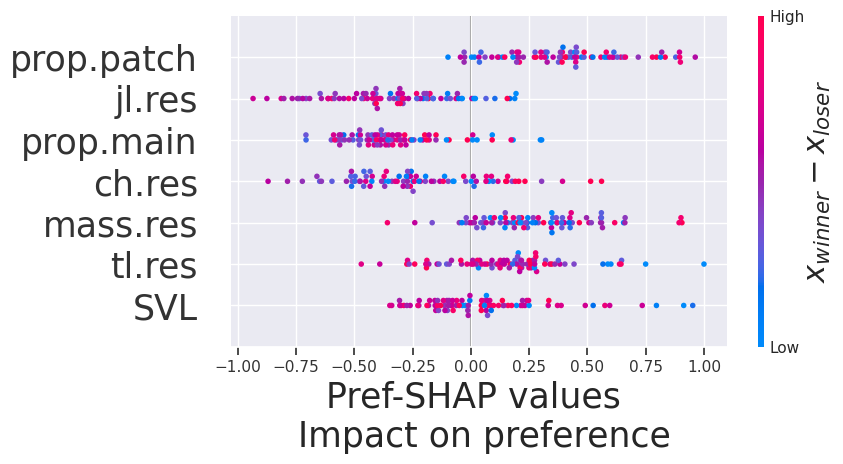}%
                \includegraphics[width=0.5\linewidth]{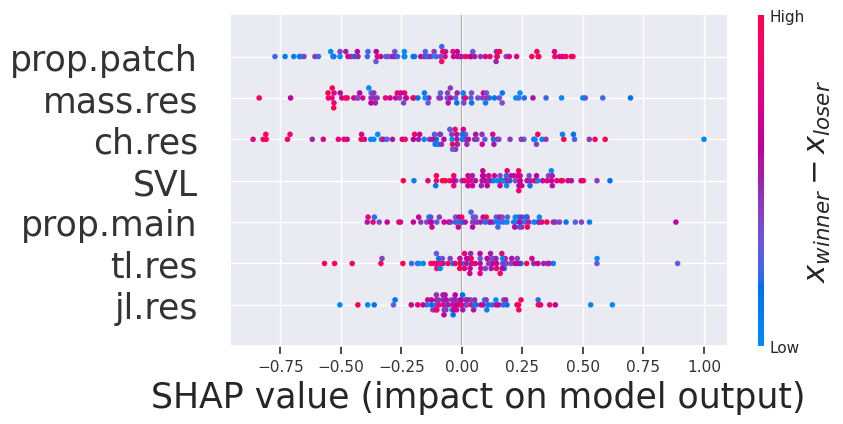}

    \end{subfigure}
         \caption{\small{Bar and Beehive plots for the Chameleon dataset}  }
  \label{chameleon_plots}
  \vspace{-0.3cm}
\end{figure}

% \paragraph{Local explanations for Pokémon battles}

\begin{figure}[htb!] %All plots between 1-5
\centering
    \begin{subfigure}[H]{0.25\linewidth}
    \centering
    \rotatebox{0}{\scalebox{0.75}{Tennis players}}
    \end{subfigure}%
    \begin{subfigure}[H]{0.25\linewidth}
    \centering
    \rotatebox{0}{\scalebox{0.75}{Environment}}
    \end{subfigure}%
   \begin{subfigure}[H]{0.25\linewidth}
    \centering
    \rotatebox{0}{\scalebox{0.75}{Tennis players}}
    \end{subfigure}%
    \begin{subfigure}[H]{0.25\linewidth}
    \centering
    \rotatebox{0}{\scalebox{0.75}{Environment}}
    \end{subfigure}
    \begin{subfigure}{0.495\textwidth}
        \centering
        \includegraphics[width=0.5\linewidth]{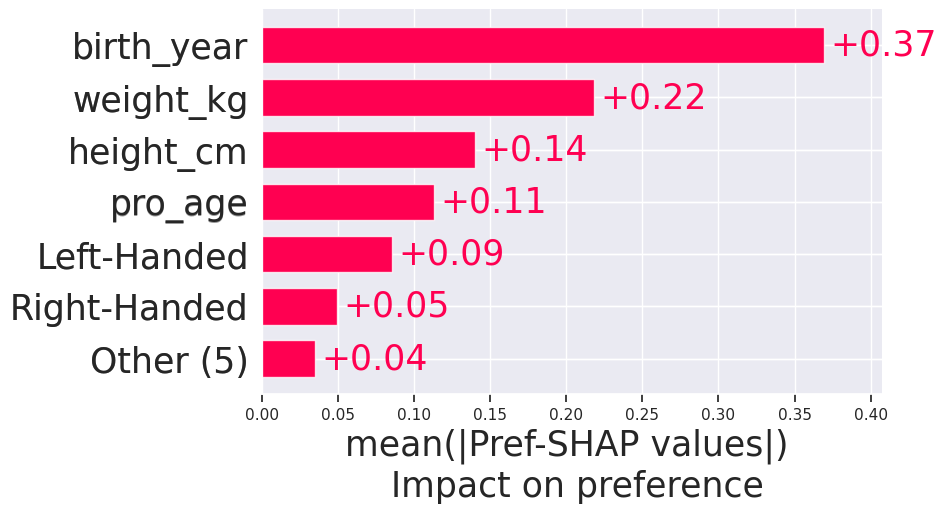}%
                \includegraphics[width=0.5\linewidth]{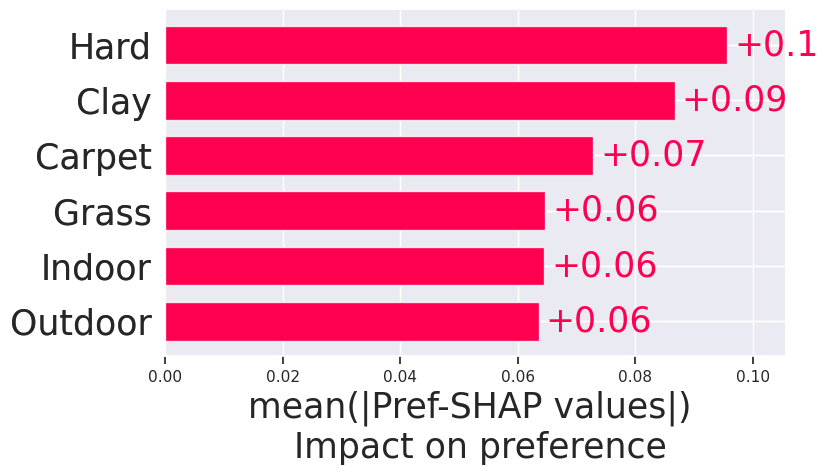}

    \end{subfigure}%
    \hfill
    \begin{subfigure}{0.495\textwidth}
        \centering
        \includegraphics[width=0.5\linewidth]{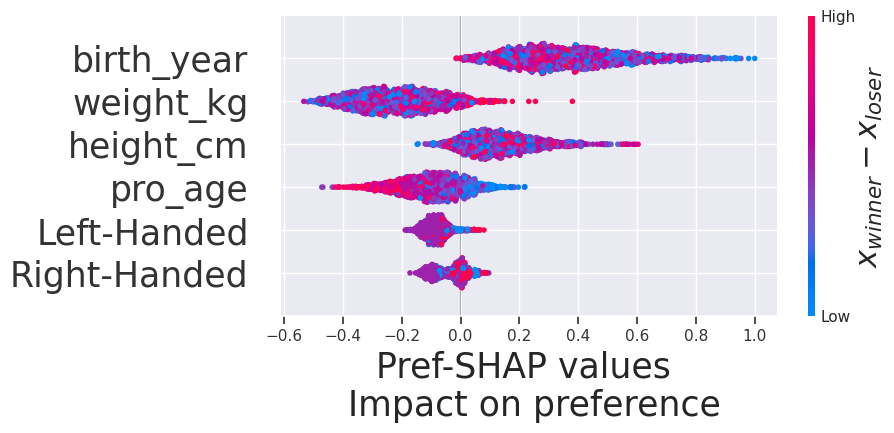}%
                \includegraphics[width=0.5\linewidth]{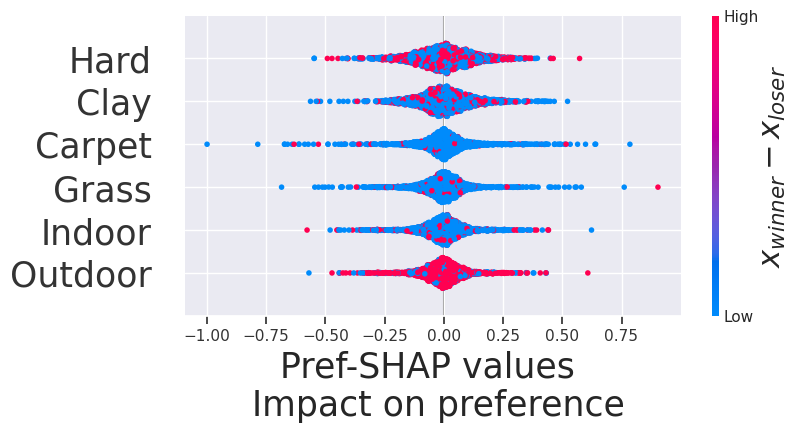}

    \end{subfigure}
            \caption{Item and context-specific Pref-SHAP values for the Tennis dataset  \\ \quad}
        \label{tennis_plots}
    \vspace{-0.5cm}
\end{figure}

\shortparagraph{Explaining Tennis matches.} We now consider preference learning with context covariates and explain both item characteristics and context covariates in \Cref{tennis_plots}. In terms of item-based inference, \textsc{Pref-SHAP} finds that being older than your opponent ($x_{\text{winner}}^{\text{yob}}-x_{\text{loser}}^{\text{yob}}<0 \to \text{Blue}$), physically heavier, and taller than your opponent positively impacts the chances of winning. We also find that debuting earlier as a professional tennis player than your opponent positively impacts your chances of winning. This is not surprising as debuting earlier may be indicative of a promising young talent. Across all competitions, there appear to be no significant patterns in environment effects.

\begin{figure}[htb!] %All plots between 1-5
\centering
    \begin{subfigure}[H]{0.25\linewidth}
    \centering
    \rotatebox{0}{\scalebox{0.75}{Djokovic}}
    \end{subfigure}%
    \begin{subfigure}[H]{0.25\linewidth}
    \centering
    \rotatebox{0}{\scalebox{0.75}{Environment}}
    \end{subfigure}%
   \begin{subfigure}[H]{0.25\linewidth}
    \centering
    \rotatebox{0}{\scalebox{0.75}{Djokovic}}
    \end{subfigure}%
    \begin{subfigure}[H]{0.25\linewidth}
    \centering
    \rotatebox{0}{\scalebox{0.75}{Environment}}
    \end{subfigure}
    \begin{subfigure}{0.495\textwidth}
        \centering
        \includegraphics[width=0.5\linewidth]{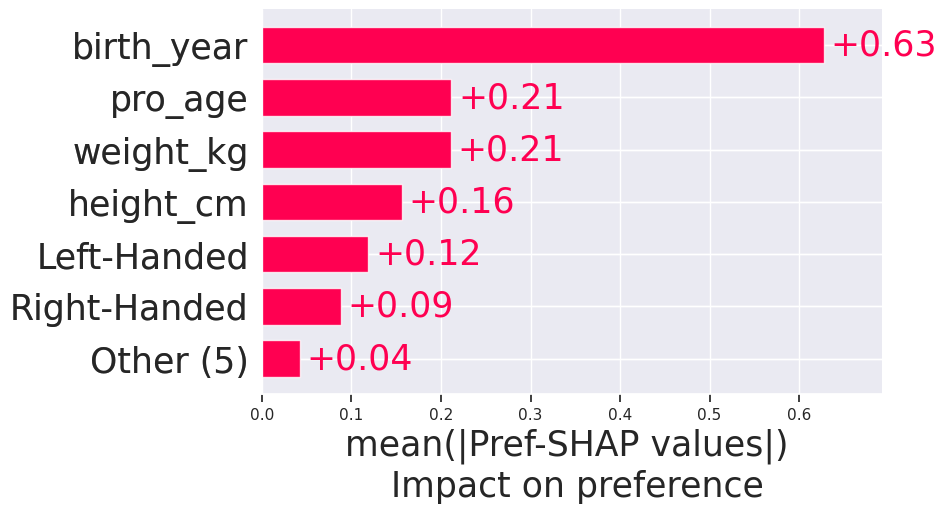}%
                \includegraphics[width=0.5\linewidth]{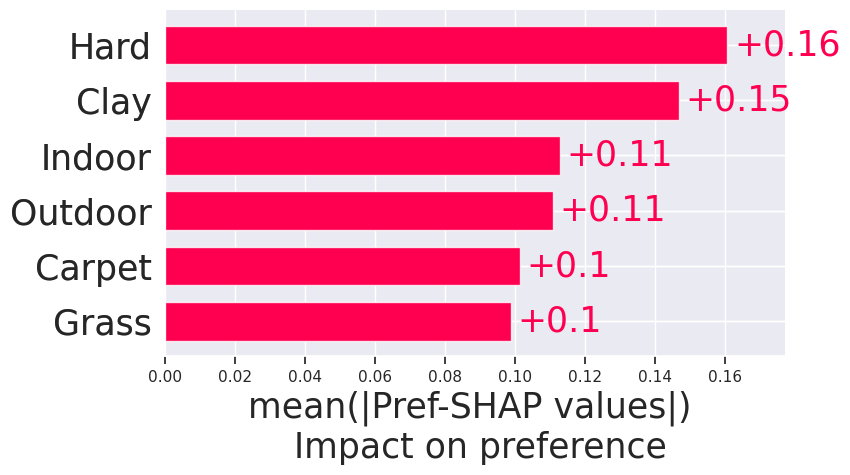}

    \end{subfigure}%
    \hfill
    \begin{subfigure}{0.495\textwidth}
        \centering
        \includegraphics[width=0.5\linewidth]{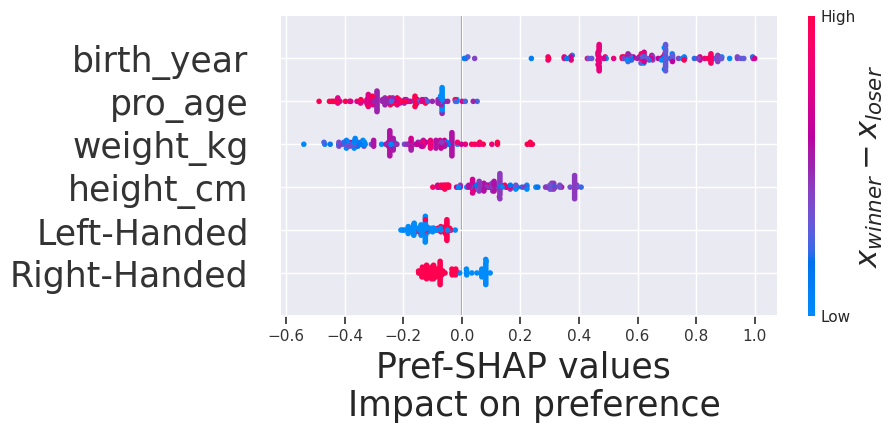}%
                \includegraphics[width=0.5\linewidth]{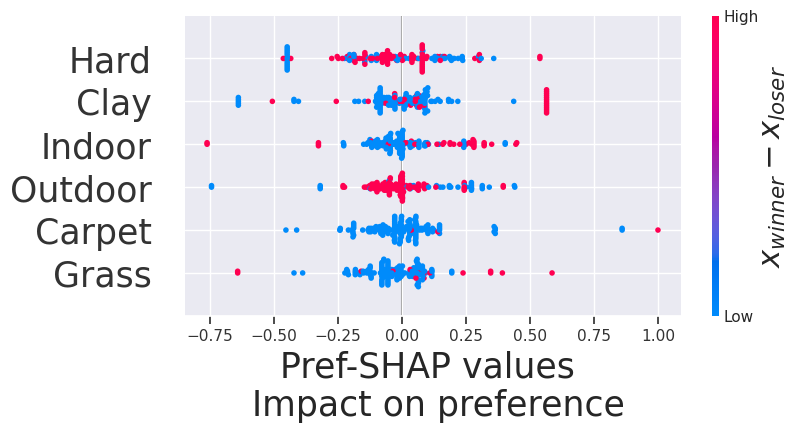}

    \end{subfigure}
            \caption{\small{Local explanations of Djokovic losses \\ \quad}}
        \label{djoko}
    \vspace{-0.7cm}
\end{figure}

\shortparagraph{Explaining Djokovic's losses}  In plot \Cref{djoko}, we locally explain all Novak Djokovic's losses in his professional career. Novak Djokovic is regarded as one of the greatest tennis players of all time, so understanding his weakness could serve as a practical demonstration of the utility of \textsc{Pref-SHAP}.

While the results take a similar shape to the global explanations, Djokovic remarkably seems to be weaker to players shorter than him, contrary to the general advantage of being taller. Besides this, Djokovic seems to be weaker on clay courts and when playing indoors. 
% From the explanations, we can infer that Djokovic has lost to items shorter and heavier than him. He is easier to beat on clay surfaces and indoors. 

\vspace{-0.2cm}
\section{Conclusion}
\vspace{-0.2cm}

In this work, we proposed \textsc{Pref-SHAP} to explain preference learning for pairwise comparison data. We proposed the appropriate value function for preference explanations and demonstrated the pathologies of the naive concatenation approach in Appendix~\ref{appendix: naive}. Experiments demonstrated that \textsc{Pref-SHAP} recovers richer explanations than utility-based approaches, showcasing the ability of \textsc{Pref-SHAP} in interpreting the mechanism of preference elicitation.

% We also demonstrated pathologies of the naive concatenation approach for preference explanation in Appendix~\ref{appendix: naive}, thus further motivation our proposed method.

% We also compared with naive concatenation approach for preference explanation in Appendix~\ref{appendix: naive} and demonstrated

% While computational optimizations have been applied to scale \textsc{Pref-SHAP}, there remains further work to truly scale to an industrial setting ($N_{\text{matches}}>10^6$). We hope that this work can be further applied and studied on safety-critical datasets in large scale settings. 

% We have proposed \textsc{Pref-SHAP} to explain \emph{inconsistent preferential learning} for both player and contest characteristics. Experiments show that \textsc{Pref-SHAP} is able to find richer interpretations than just explaining UPM, demonstrating the importance of interpretable general preferential learning for better understanding of duelling data. While computational optimizations have been applied to scale \textsc{Pref-SHAP}, there remains further work to truly scale to an industrial setting ($N_{\text{matches}}>10^6$). We hope that this work can be further applied and studied on safety-critical datasets in large scale settings. 

\ack
The authors sincerely thank Oscar Clivio for the helpful comments and feedback. SLC is supported by the EPSRC and MRC through the OxWaSP CDT programme EP/L016710/1. DS is supported in part by Tencent AI Lab and in part by the Alan Turing Institute (EP/N510129/1).

\clearpage
\bibliographystyle{unsrtnat}
\bibliography{main}

\clearpage

%%%%%%%%%%%%%%%%%%%%%%%%%%%%%%%%%%%%%%%%%%%%%%%%%%%%%%%%%%%
\section*{Checklist}

\begin{enumerate}

\item For all authors...
\begin{enumerate}
  \item Do the main claims made in the abstract and introduction accurately reflect the paper's contributions and scope?
    \answerYes{}
    \item Did you describe the limitations of your work?
  \answerYes{}
  \item Did you discuss any potential negative societal impacts of your work?
    \answerNA{}
  \item Have you read the ethics review guidelines and ensured that your paper conforms to them?
    \answerYes{}
\end{enumerate}

\item If you are including theoretical results...
\begin{enumerate}
  \item Did you state the full set of assumptions of all theoretical results?
    \answerYes{}
        \item Did you include complete proofs of all theoretical results?
    \answerYes{}
\end{enumerate}

\item If you ran experiments...
\begin{enumerate}
  \item Did you include the code, data, and instructions needed to reproduce the main experimental results (either in the supplemental material or as a URL)?
    \answerYes{}
  \item Did you specify all the training details (e.g., data splits, hyperparameters, how they were chosen)?
    \answerYes{}
        \item Did you report error bars (e.g., with respect to the random seed after running experiments multiple times)?
    \answerNA{}
        \item Did you include the total amount of compute and the type of resources used (e.g., type of GPUs, internal cluster, or cloud provider)? See Appendix \ref{compute_stuff}
    \answerYes{}
\end{enumerate}

\item If you are using existing assets (e.g., code, data, models) or curating/releasing new assets...
\begin{enumerate}
  \item If your work uses existing assets, did you cite the creators?
    \answerYes{}
  \item Did you mention the license of the assets?
    \answerNA{}
  \item Did you include any new assets either in the supplemental material or as a URL?
    \answerNA{}
  \item Did you discuss whether and how consent was obtained from people whose data you're using/curating?
    \answerNA{}
  \item Did you discuss whether the data you are using/curating contains personally identifiable information or offensive content?
    \answerNA{}
\end{enumerate}

\item If you used crowdsourcing or conducted research with human subjects...
\begin{enumerate}
  \item Did you include the full text of instructions given to participants and screenshots, if applicable?
    \answerNA{}
  \item Did you describe any potential participant risks, with links to Institutional Review Board (IRB) approvals, if applicable?
    \answerNA{}
  \item Did you include the estimated hourly wage paid to participants and the total amount spent on participant compensation?
    \answerNA{}
\end{enumerate}

\end{enumerate}

%%%%%%%%%%%%%%%%%%%%%%%%%%%%%%%%%%%%%%%%%%%%%%%%%%%%%%%%%%%%
\newpage

\appendix

\section{Computation and Implementation Details}
\label{compute_stuff}

We propose several optimizations in the \textsc{Pref-SHAP} procedure. We consider fast sampling of coalitions $S$ in Algorithm \ref{algo1} batched conjugate gradient descent in Algorithm \ref{algo2} described below. \newline 
\textbf{Fast coalitions} \quad We first propose an optimized sampling scheme for finding coalitions $S$ in Algorithm \ref{algo1}.
\begin{algorithm}[htb!]
\caption{Sampling unique $n$, $d$-dimensional coalitions in $\mathcal{O}(d)$ time}
\label{algo1}
\begin{algorithmic}[1]
\Require Number of coalitions $n$, number of features $d$.
\State $I=\text{SampleWithoutReplacement}(n,0,2^d)$ \Comment{Sample $n$ unique integers between $0$ and $2^d$}
\Function{\textnormal{base2}}{i}: \Comment{Convert integer to base-2 representation}
\State $S = [0,\hdots, 0] \in \mathbb{R}^d$, $r=i$ \Comment{Initialize $d$-dimensional 0 vector and the rest term $r$}
\While {$r>0$}
\State $i = \lfloor\log_2{(r)}\rfloor$ \Comment{Find which index of $S$ to set to 1}
\State $S[i]=1$ \Comment{Update $S$}
\State $r = r -2^{i}$ \Comment{Update rest term}
\EndWhile
\State \textbf{return} $S$
\EndFunction %
\Return $\{S_1,\hdots S_{n}\}=\text{parallel\_apply}(I,\text{base2})$ \Comment{Each integer can be independently converted}
\end{algorithmic}
\end{algorithm}
In contrast to the implementation in \cite{lundberg2017unified} which samples the weights from $p(Z)$, our method is embarrassingly parallel, which allows for an additional $\mathcal{O}(n)$ reduction. A naive algorithm that compares each sample $S_i$ has complexity $\mathcal{O}(n^2d^2)$ and cannot be parallelized.\newline  \textbf{Stabilizing the Shapley value estimation}\quad
We remove the features which have $0$ variance in the data we are explaining, similar to the implementation in SHAP. %For the coalition $S$, we simply set filtered out features to $0$. 
To ensure we get numerically stable Shapley Values, we calculate the inverse using Cholesky decomposition, as we found the regular inverse function provided inconsistent results.  

To calculate CMEs effectively, we use  \emph{preconditioned batched} conjugate gradient descent over coalitions detailed in Algorithm \ref{algo2}.
\begin{algorithm}[H]
\caption{Batched conjugate gradient descent}
\label{algo2}
\begin{algorithmic}
\Require Preconditioner $P=\mathbf{K}_{\mathbf{x}_D}^{-1}$, batch $\mathbf{X} = [\mathbf{K}_{\mathbf{x}_{S_i}} \hdots \mathbf{K}_{\mathbf{x}_{S_n}}]$, $\mathbf{B} = [\mathbf{K}_{\mathbf{x}_{S_i},x_{S_i}}\hdots \mathbf{K}_{\mathbf{x}_{S_n},x_{S_n}}]$,  \texttt{max\_its}, tolerance $\varepsilon$
\State Set $\mathbf{R} = \mathbf{B}$, $\mathbf{Z} = \text{BatchMM}(P,\mathbf{B}), \mathbf{p} = \mathbf{Z}$, $\mathbf{a}=\mathbf{0}$
\State Set $\mathbf{R}_Z = [(\mathbf{R}_i\circ \mathbf{Z}_i)_{++} \hdots (\mathbf{R}_n\circ \mathbf{Z}_n)_{++}] $ \Comment{Element wise product and sum}
\For { \texttt{max\_its}}
\State $\mathbf{L}= \text{BatchMM}(\mathbf{X},\mathbf{B})$
\State $\boldsymbol{\alpha}= \mathbf{L}\circ \frac{1}{[(\mathbf{L}_i\circ \mathbf{p}_i)_{++} \hdots (\mathbf{L}_n\circ \mathbf{p}_n)_{++}]}$
\State $\mathbf{a} = \mathbf{a} +\boldsymbol{\alpha} \circ \mathbf{p}$
\State $\mathbf{r} = \mathbf{r} -\boldsymbol{\alpha} \circ \mathbf{L}$
\If {Mean($[(\mathbf{r}_i\circ \mathbf{r}_i)_{++} \hdots (\mathbf{r}_n\circ \mathbf{r}_n)_{++}]$) < $\varepsilon$}
\Return $\mathbf{a}$
\EndIf
\State $\mathbf{z} = \text{BatchMM}(P,\mathbf{r})$
\State $\mathbf{R}_Z^{\text{new}} = [(\mathbf{R}_i\circ \mathbf{Z}_i)_{++} \hdots (\mathbf{R}_n\circ \mathbf{Z}_n)_{++}] $
\State $\mathbf{p}=\mathbf{z}+[(\mathbf{R}_i^{\text{new}}\circ \frac{1}{\mathbf{R}_i})_{++} \hdots (\mathbf{R}_n^{\text{new}}\circ \frac{1}{\mathbf{R}_n})_{++}] \circ \mathbf{p}$
\State $\mathbf{R}_Z=\mathbf{R}_Z^{\text{new}}$
\EndFor \\
\Return $\mathbf{a}$
\end{algorithmic}
\end{algorithm}

We have run all our jobs on one Nvidia V100 GPU.

\section{Additional Experimental Results}
\label{appendix: naive}
\shortparagraph{Naive Concatenation} We demonstrate the pathologies of the naive concatenation approach mentioned in Sec. 3 with our synthetic experiment. Recall that naive-concatenation approach here corresponds to first concatenating $\bfxleft,\bfxright$'s features together and applying SHAP to the learned function $g$ directly, in order to to obtain $2d$ Shapley values, instead of the original $d$, since each feature has been duplicated. This approach ignores that the items $\bfxleft$ and $\bfxright$ in fact consist of the same features. Therefore, when we use the usual value function from SHAP (corresponding to the impact an individual feature has on the model when it is turned ``off'' by integration), we would be turning ``off'' the feature from the left item, while keeping ``on'' the feature from the right item, obtaining a difficult to interpret attribution score. This is highly problematic, as we might be inferring vastly different contributions of the same feature purely because of the item ordering when concatenating them. We note that the item ordering in all our experiments is arbitrary and carries no additional information about the match.%Why? Recall that value functions are defined intuitively as the impact one feature has on the model when it is turned ''off'' by integration. %Now with this duplicating approach, we are ignoring the fact that features from $\bfxleft,\bfxright$ are actually the same, and thus obtain unreasonable "utility" scores. 

\begin{figure}[htb!!!]
  \begin{center}
    \includegraphics[width=0.5\textwidth]{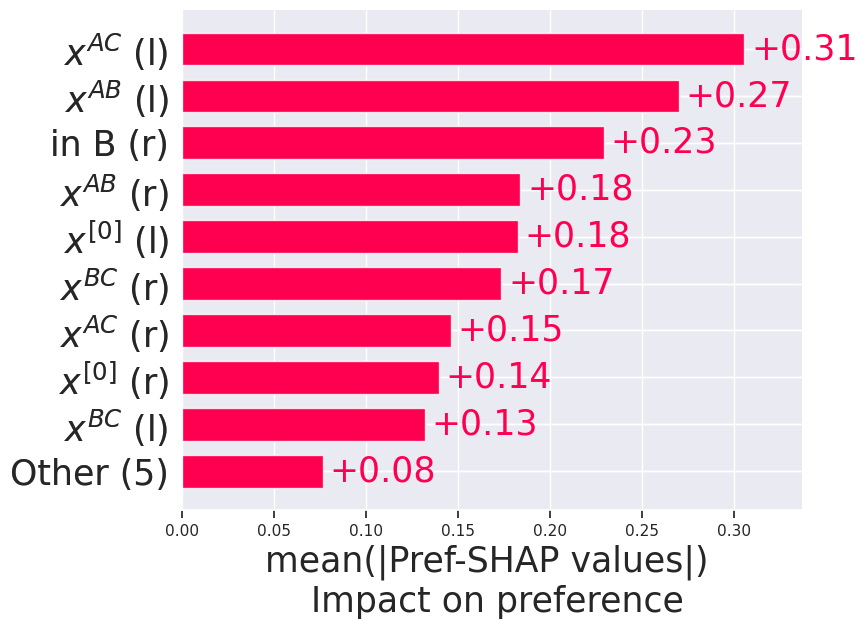}%
        \includegraphics[width=0.5\textwidth]{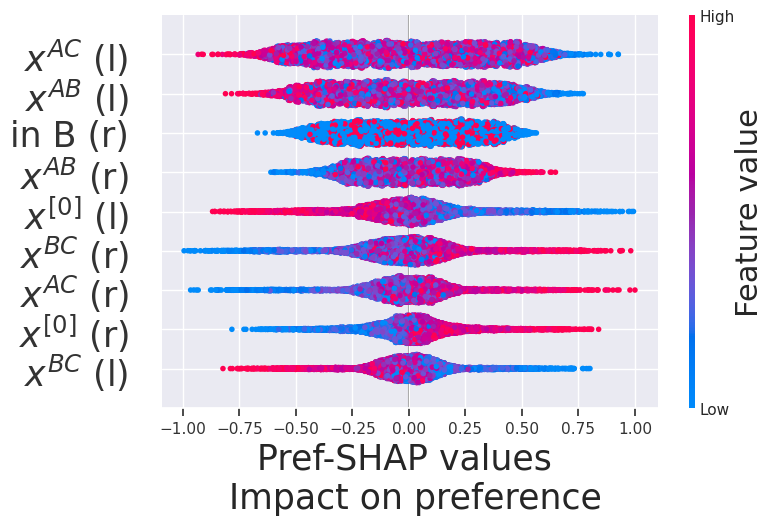}
  \end{center}
  \caption{Explaining a naïve concatenation model}
  \label{naive}
\end{figure}

We can see from Fig.~\ref{naive} that when we explain the preference model applied to the synthetic experiment, we see that, for example, $x^{AC}(l)$ from $\bfxleft$ and $x^{AC}(r)$ from $\bfxright$ have in fact very different average Shapley values. Even attempting to average each pair of corresponding features does not give a meaningful feature contribution ordering ($x^{AB}$ and $x^{AC}$ are scored higher on average than $x^{BC}$ and $x^{[0]}$). 
% The test AUC of the Naive concatenation model is $0.964 \pm0.001$. The GPM model which achieved $0.981 \pm0.002$.  

%where in fact there is no clear preference between $\bfxleft,\bfxright$ in our experiments.

% \paragraph{Summing or subtracting Shapley values do not recover meaningful insights} To obtain a summary of contributions for each feature items from the concatenation approach, we add~(Fig.~\ref{naive_add}) and subtract~(Fig.~\ref{naive_sub}) the Shapley values obtained from left and right players, respectively. We see  

% \shortparagraph{Summing and subtracting Shapley values} We further consider both subtracting the and adding the Shapley values with $\bfx^{\text{winner}}-\bfx^{\text{loser}}$ and $\bfx^{\text{winner}}+\bfx^{\text{loser}}$ respectively, to obtain a summary of item feature contributions. We show these plots in \Cref{naive_sub} and \Cref{naive_add}. We see that both subtracting and adding does not recover the same insight as \textsc{Pref-SHAP}. The addition recovers the correct absolute magnitude of the Shapley values, however the beeplot does not reflect the correct insights. 

% \begin{figure}[htb!!!]
%   \begin{center}
%     \includegraphics[width=0.5\textwidth]{subtract_addition/bar_plot_lamb=0.0_lasso.png}%
%         \includegraphics[width=0.5\textwidth]{subtract_addition/bee_plot_lamb=0.0_lasso.png}
%   \end{center}
%   \caption{Explaining a naïve concatenation model, subtracting}
%   \label{naive_sub}
% \end{figure}

% \begin{figure}[htb!!!]
%   \begin{center}
%     \includegraphics[width=0.5\textwidth]{addition_break/bar_plot_lamb=0.0_lasso.png}%
%         \includegraphics[width=0.5\textwidth]{addition_break/bee_plot_lamb=0.0_lasso.png}
%   \end{center}
%   \caption{Explaining a naïve concatenation model, adding}
%   \label{naive_add}
% \end{figure}

\paragraph{Additional synthetic data}
We consider an additional synthetic experiment where we generate data directly from a GPM model and one where we construct synthetic dueling data. When simulating data, we first generate player covariates as  $\bfx_i \in \mathbb{R}^d \sim \mathcal{N}(0,0.1\textbf{I}_d)$ for each player $i$. When generating from the GPM model, we would set 2 covariates as important, by only keeping the 2 first entries of $\bfx_i$ and fixing the rest to be constant (equal to 0). We build a GPM model for $g$ out of these covariates and generate match outcomes.
% When generating data directly from a GPM model, we first simulate covariates $\bfx_i \in \mathbb{R}^d \sim \mathcal{N}(0,0.1\textbf{I}_d)$ for each player $i$. If we want $d'$ specific covariates to be important, we fix the remaining covariates $d-d'$ to be constant ones. We build a GPM model out of these covariates and generate winners and loosers.

We consider $d=10$, where only the two first features are set to be important in predicting the outcome.
\begin{figure}[htb!] %All plots between 1-5

\centering
    \begin{subfigure}[H]{0.25\linewidth}
    \centering
    \rotatebox{0}{\scalebox{0.75}{\textsc{Pref-SHAP}}}
    \end{subfigure}%
    \begin{subfigure}[H]{0.25\linewidth}
    \centering
    \rotatebox{0}{\scalebox{0.75}{Explaining UPM}}
    \end{subfigure}%
   \begin{subfigure}[H]{0.25\linewidth}
    \centering
    \rotatebox{0}{\scalebox{0.75}{\textsc{Pref-SHAP}}}
    \end{subfigure}%
    \begin{subfigure}[H]{0.25\linewidth}
    \centering
    \rotatebox{0}{\scalebox{0.75}{Explaining UPM}}
    \end{subfigure}
    \begin{subfigure}{0.495\textwidth}
        \centering
        \includegraphics[width=0.5\linewidth]{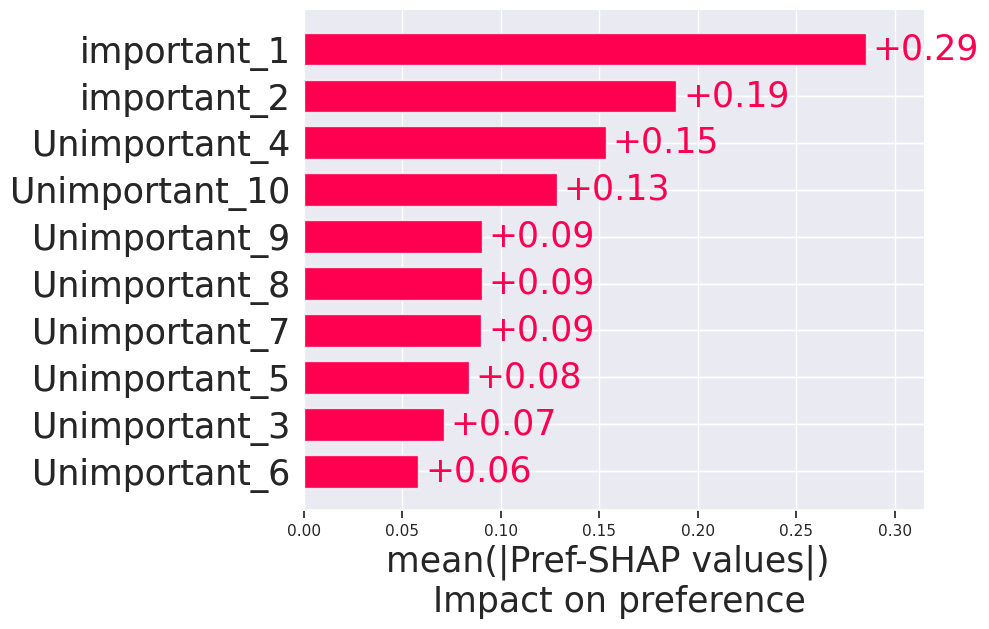}%
                \includegraphics[width=0.5\linewidth]{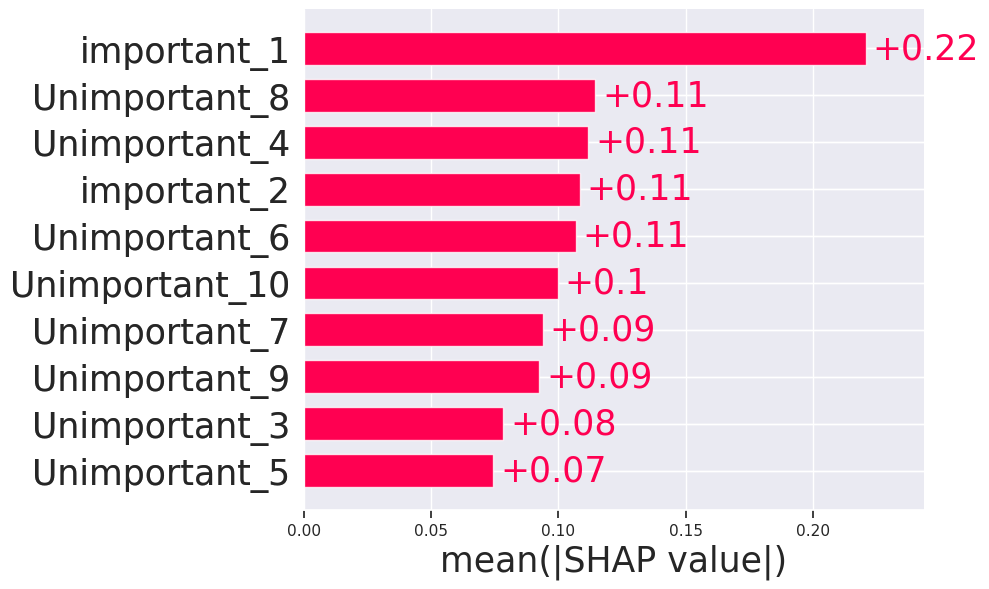}
        % \includegraphics[width=\linewidth]{test/bee_plot_lamb=0.0_ridge.png}
        % \caption{Barplots for the Pokémon dataset. Shapley values calculated on test data. \\ \quad}
        % \label{synt_barplot}
    \end{subfigure}%
    \hfill
    \begin{subfigure}{0.495\textwidth}
        \centering
        \includegraphics[width=0.5\linewidth]{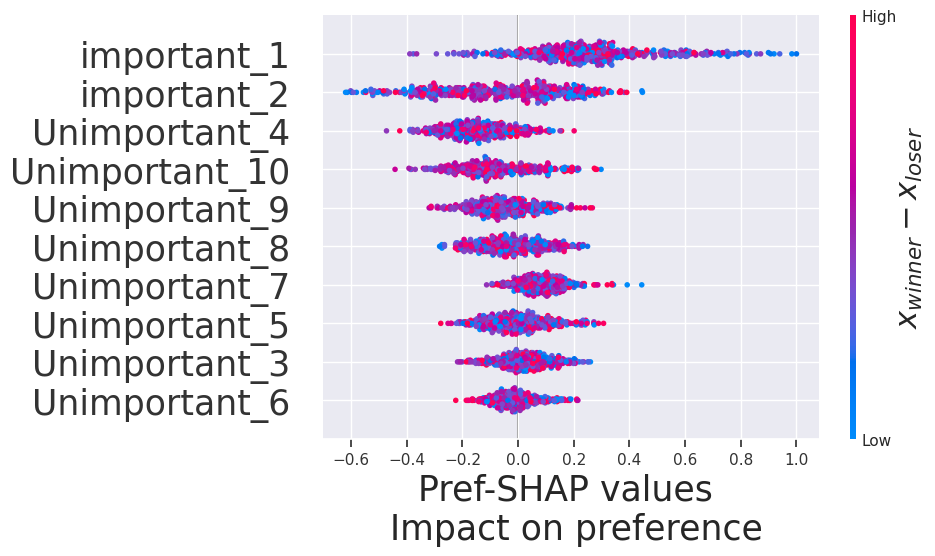}%
                \includegraphics[width=0.5\linewidth]{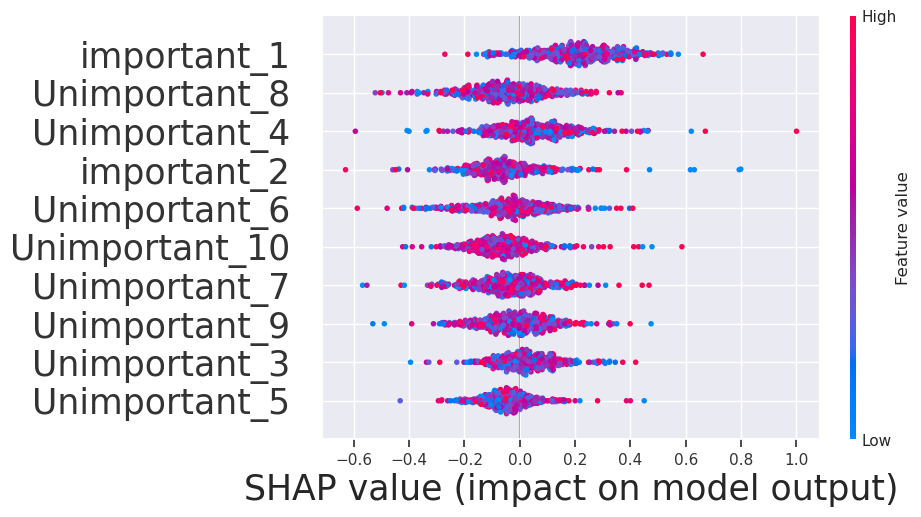}
        % \caption{Beehive plots for the Pokémon dataset. \textsc{Pref-SHAP} captures that both speed and type matter, while explaining rank only captures the type importance.}
        % \label{synt_beeplot}
    \end{subfigure}
     \caption{Bar and Beehive plots for Simulation A. \textsc{Pref-SHAP} recovers the correct features (1,2), while explaining UPM does not.}
  \label{synt_beeplot}
\end{figure}

\begin{figure}[htb!] %All plots between 1-5

\centering
\begin{subfigure}[H]{0.33\linewidth}
    \centering
    \includegraphics[width=\linewidth]{chameleon_gpgp/bee_plot_lamb=0.0_lasso.png}
\end{subfigure}%
\begin{subfigure}[H]{0.33\linewidth}
    \centering
    \includegraphics[width=\linewidth]{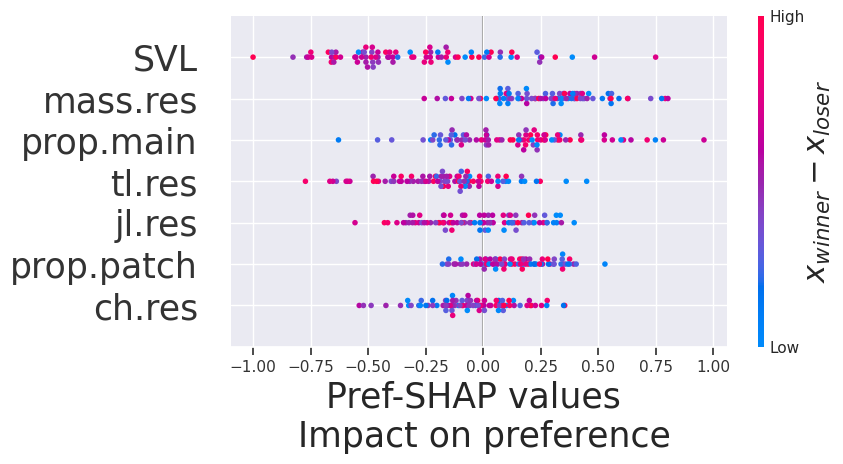}
\end{subfigure}%
\begin{subfigure}[H]{0.33\linewidth}
    \centering
    \includegraphics[width=\linewidth]{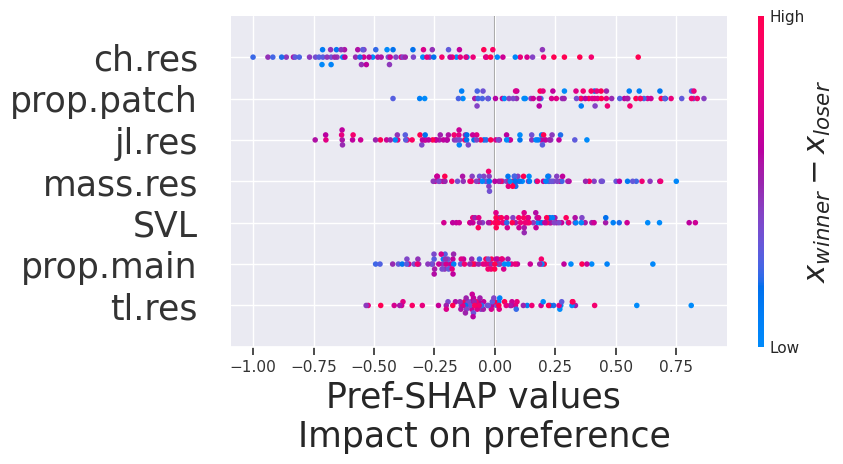}
\end{subfigure}
\begin{subfigure}[H]{0.33\linewidth}
    \centering
    \includegraphics[width=\linewidth]{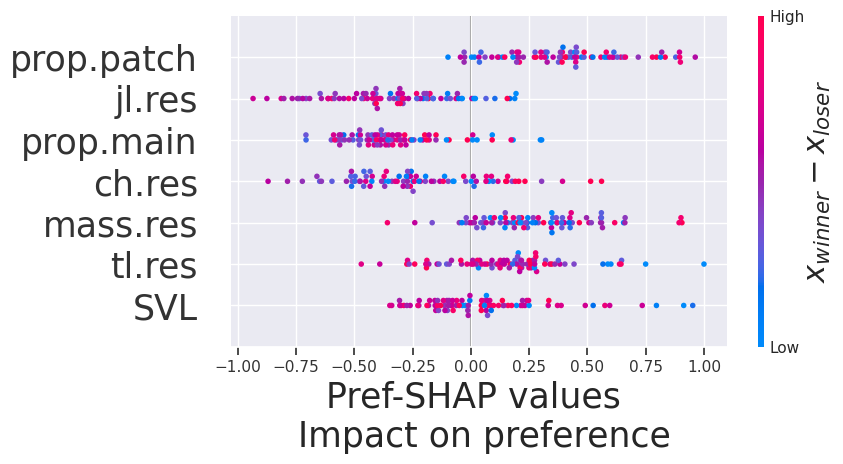}
\end{subfigure}%
\begin{subfigure}[H]{0.33\linewidth}
    \centering
    \includegraphics[width=\linewidth]{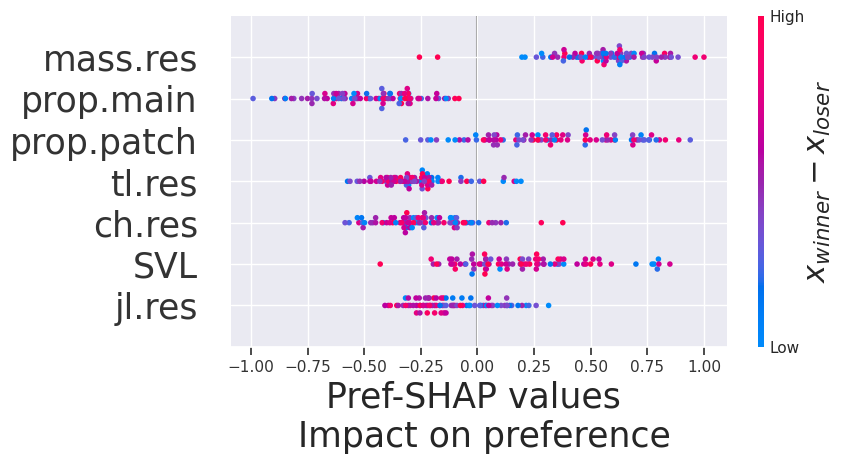}
\end{subfigure}%
\begin{subfigure}[H]{0.33\linewidth}
    \centering
    \includegraphics[width=\linewidth]{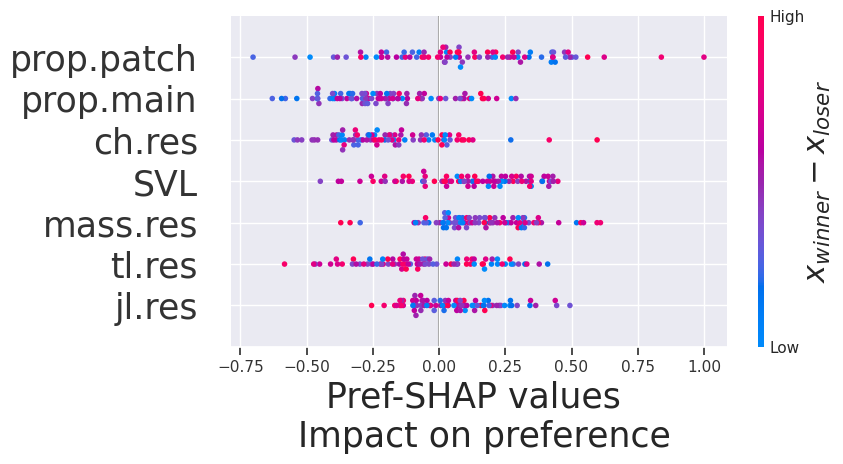}
\end{subfigure}
\caption{\textsc{Pref-SHAP} applied to 6 different folds of Chameleon. \textsc{Pref-SHAP} consistently finds that higher values of \textit{jaw length} (jl.res) have a negative impact on outcome.}
\label{cham_app}
\end{figure}

\begin{figure}[htb!] %All plots between 1-5

\centering
\begin{subfigure}[H]{0.33\linewidth}
    \centering
    \includegraphics[width=\linewidth]{chameleon_pgp/bee_plot_lamb=0.0_lasso.png}
\end{subfigure}%
\begin{subfigure}[H]{0.33\linewidth}
    \centering
    \includegraphics[width=\linewidth]{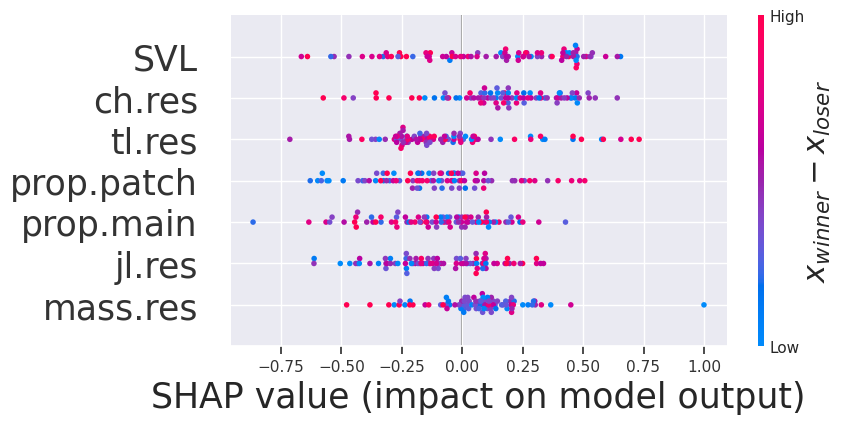}
\end{subfigure}%
\begin{subfigure}[H]{0.33\linewidth}
    \centering
    \includegraphics[width=\linewidth]{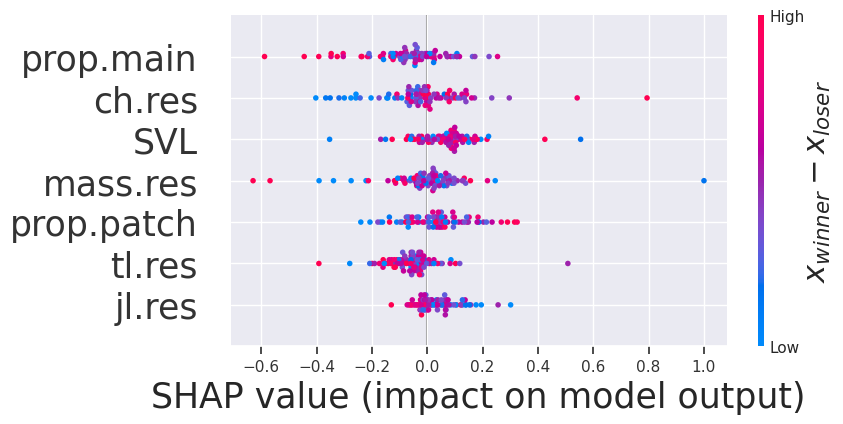}
\end{subfigure}
\begin{subfigure}[H]{0.33\linewidth}
    \centering
    \includegraphics[width=\linewidth]{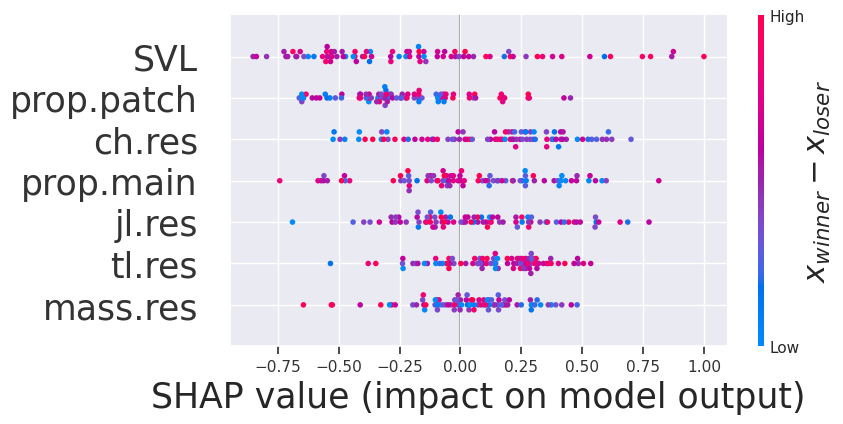}
\end{subfigure}%
\begin{subfigure}[H]{0.33\linewidth}
    \centering
    \includegraphics[width=\linewidth]{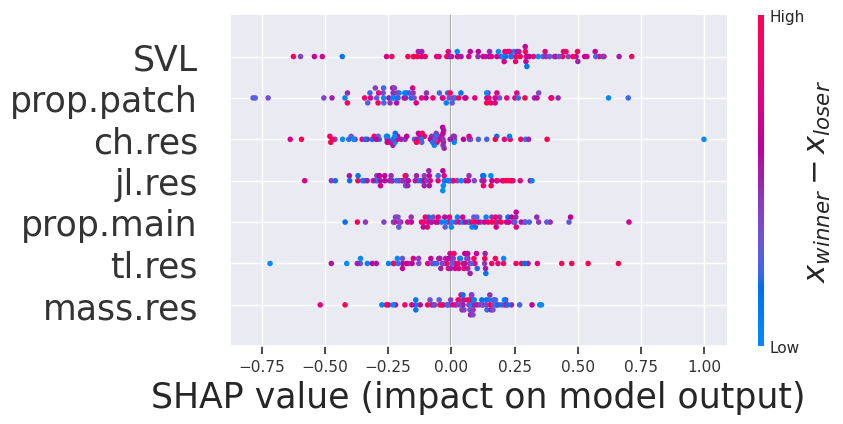}
\end{subfigure}%
\begin{subfigure}[H]{0.33\linewidth}
    \centering
    \includegraphics[width=\linewidth]{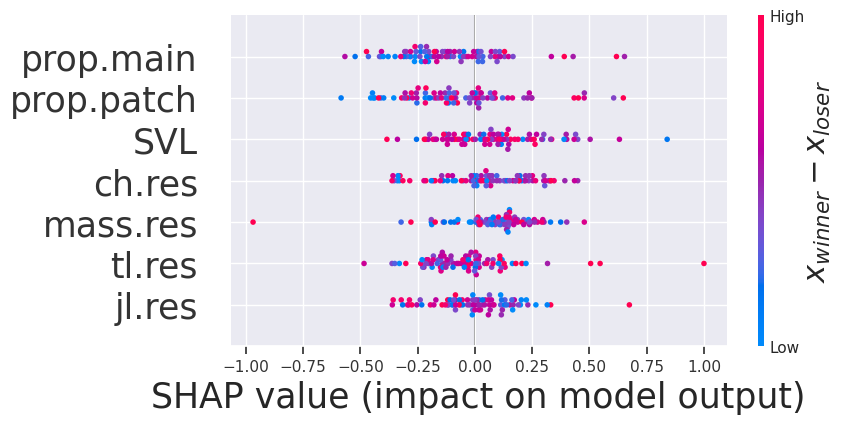}
\end{subfigure}
\caption{UPM explanations on 6 different folds of Chameleon. UPM is unable to find a consistent pattern for the impact of \textit{jaw length} (jl.res) on the outcome.}
\label{cham_app_upm}

\end{figure}

\paragraph{Chameleon data}
\label{cham_proof}
We further provide explanations of the Chameleon dataset on several different folds in \Cref{cham_app} and \Cref{cham_app_upm}.

\paragraph{Additional local explanations}
\label{local_exp_appendix}

\begin{figure}[htb!] %All plots between 1-5

\centering
    \begin{subfigure}[H]{0.25\linewidth}
    \centering
    \rotatebox{0}{\scalebox{0.75}{\textsc{Pref-SHAP}}}
    \end{subfigure}%
    \begin{subfigure}[H]{0.25\linewidth}
    \centering
    \rotatebox{0}{\scalebox{0.75}{Explaining UPM}}
    \end{subfigure}%
   \begin{subfigure}[H]{0.25\linewidth}
    \centering
    \rotatebox{0}{\scalebox{0.75}{\textsc{Pref-SHAP}}}
    \end{subfigure}%
    \begin{subfigure}[H]{0.25\linewidth}
    \centering
    \rotatebox{0}{\scalebox{0.75}{Explaining UPM}}
    \end{subfigure}
    \begin{subfigure}{0.495\textwidth}
        \centering
        \includegraphics[width=0.5\linewidth]{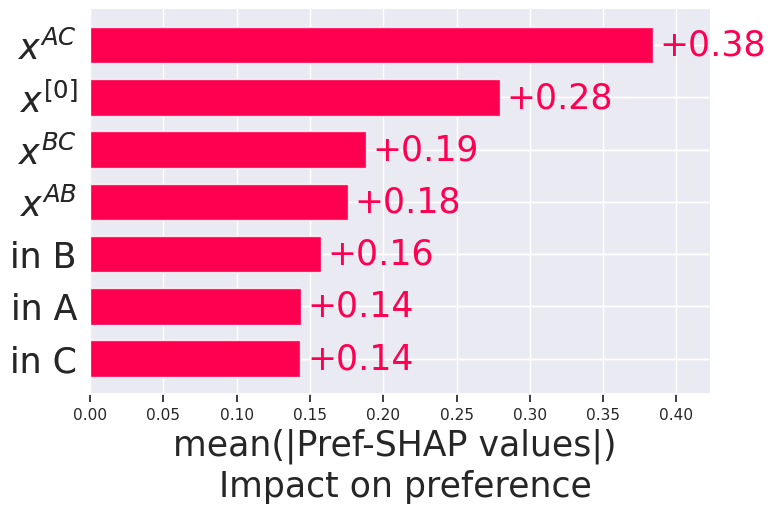}%
                \includegraphics[width=0.5\linewidth]{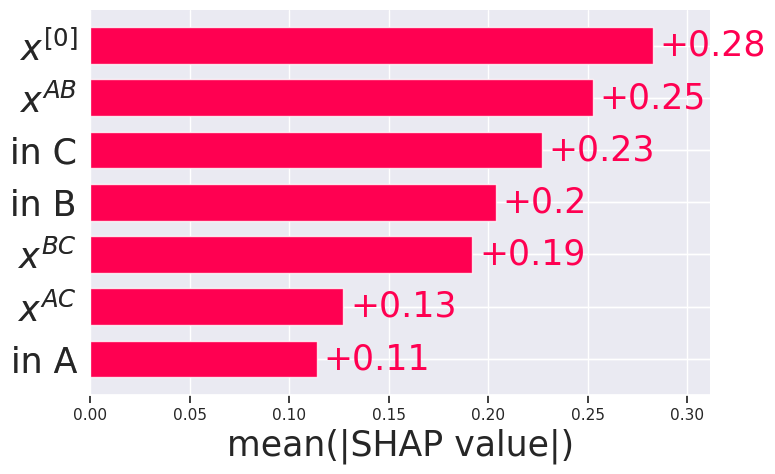}
        % \includegraphics[width=\linewidth]{test/bee_plot_lamb=0.0_ridge.png}
        % \caption{Barplots for the Pokémon dataset. Shapley values calculated on test data. \\ \quad}
        % \label{synt_barplot}
    \end{subfigure}%
    \hfill
    \begin{subfigure}{0.495\textwidth}
        \centering
        \includegraphics[width=0.5\linewidth]{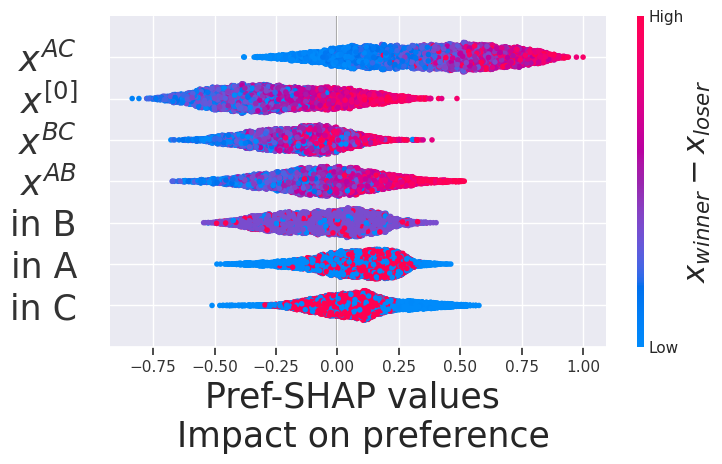}%
                \includegraphics[width=0.5\linewidth]{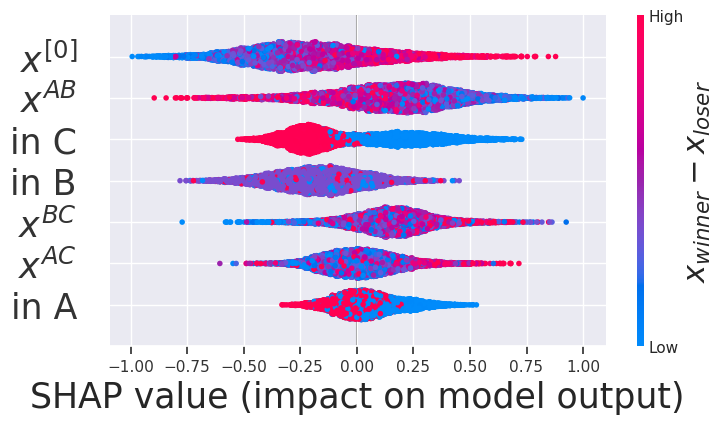}
        % \caption{Beehive plots for the Pokémon dataset. \textsc{Pref-SHAP} captures that both speed and type matter, while explaining rank only captures the type importance.}
        % \label{synt_beeplot}
    \end{subfigure}
     \caption{Explaining matches between clusters 0 and 2 on the synthetic dataset.}
  \label{synt_beeplot_appendix_1}
\end{figure}

\begin{figure}[htb!] %All plots between 1-5

\centering
    \begin{subfigure}[H]{0.25\linewidth}
    \centering
    \rotatebox{0}{\scalebox{0.75}{\textsc{Pref-SHAP}}}
    \end{subfigure}%
    \begin{subfigure}[H]{0.25\linewidth}
    \centering
    \rotatebox{0}{\scalebox{0.75}{Explaining UPM}}
    \end{subfigure}%
   \begin{subfigure}[H]{0.25\linewidth}
    \centering
    \rotatebox{0}{\scalebox{0.75}{\textsc{Pref-SHAP}}}
    \end{subfigure}%
    \begin{subfigure}[H]{0.25\linewidth}
    \centering
    \rotatebox{0}{\scalebox{0.75}{Explaining UPM}}
    \end{subfigure}
    \begin{subfigure}{0.495\textwidth}
        \centering
        \includegraphics[width=0.5\linewidth]{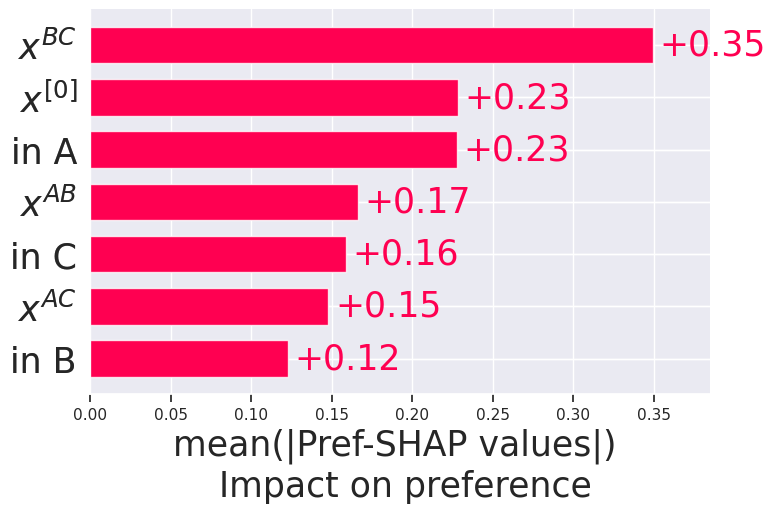}%
                \includegraphics[width=0.5\linewidth]{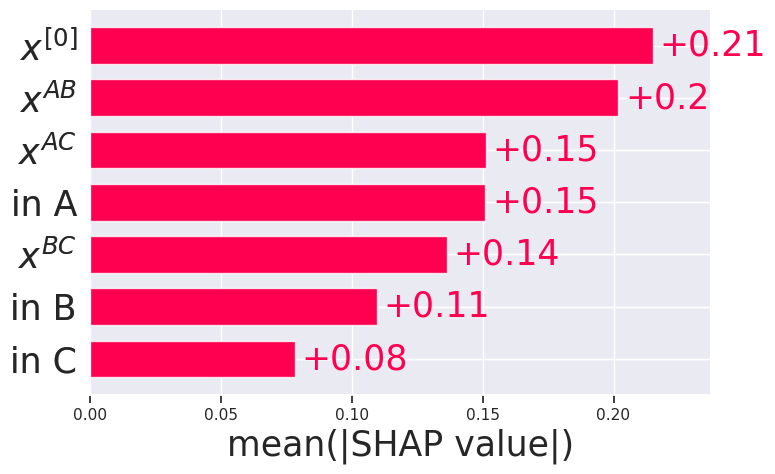}
        % \includegraphics[width=\linewidth]{test/bee_plot_lamb=0.0_ridge.png}
        % \caption{Barplots for the Pokémon dataset. Shapley values calculated on test data. \\ \quad}
        % \label{synt_barplot}
    \end{subfigure}%
    \hfill
    \begin{subfigure}{0.495\textwidth}
        \centering
        \includegraphics[width=0.5\linewidth]{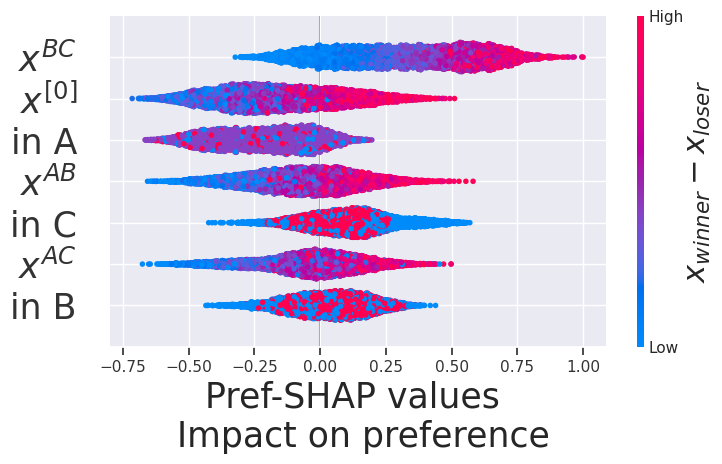}%
                \includegraphics[width=0.5\linewidth]{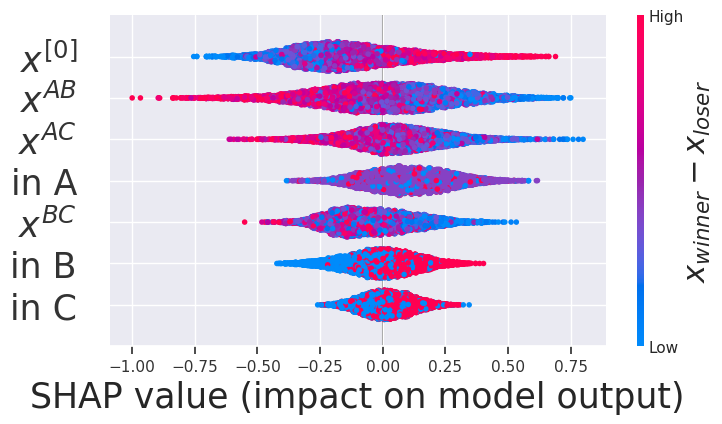}
        % \caption{Beehive plots for the Pokémon dataset. \textsc{Pref-SHAP} captures that both speed and type matter, while explaining rank only captures the type importance.}
        % \label{synt_beeplot}
    \end{subfigure}
     \caption{Explaining matches between clusters 1 and 2 on the synthetic dataset.}
  \label{synt_beeplot_appendix_2}
\end{figure}

We additionally provide local explanations on the synthetic dataset in \Cref{synt_beeplot_appendix_1} and \Cref{synt_beeplot_appendix_2}.

\paragraph{Website dataset}
\label{website_stuff}
\textit{The Website} dataset considers anonymized visitors on a fashion retail website, where we are given what garment each visitor viewed and what each visitor clicked in a session. A user may have more than one session. In this setup, we interpret a browsing session for a visitor as multiple matches between items, such that the winning item (clicked) competes against all losing items (only viewed). If several items are winners, they do not play against each other. Each item has several descriptive statistics such as \textit{colour}, \textit{garment type}, \textit{assortment characteristic} etc. There are some limited descriptive statistics of the visitors, such as \textit{year of birth} and \textit{gender code} (i.e. Male/Female/Unspecified/Unknown). 

\begin{figure}[H] %All plots between 1-5
\centering
    \begin{subfigure}[H]{0.25\linewidth}
    \centering
    \rotatebox{0}{\scalebox{0.75}{Products}}
    \end{subfigure}%
    \begin{subfigure}[H]{0.25\linewidth}
    \centering
    \rotatebox{0}{\scalebox{0.75}{User}}
    \end{subfigure}%
   \begin{subfigure}[H]{0.25\linewidth}
    \centering
    \rotatebox{0}{\scalebox{0.75}{Products}}
    \end{subfigure}%
    \begin{subfigure}[H]{0.25\linewidth}
    \centering
    \rotatebox{0}{\scalebox{0.75}{User}}
    \end{subfigure}
    \begin{subfigure}{0.495\textwidth}
        \centering
        \includegraphics[width=0.5\linewidth]{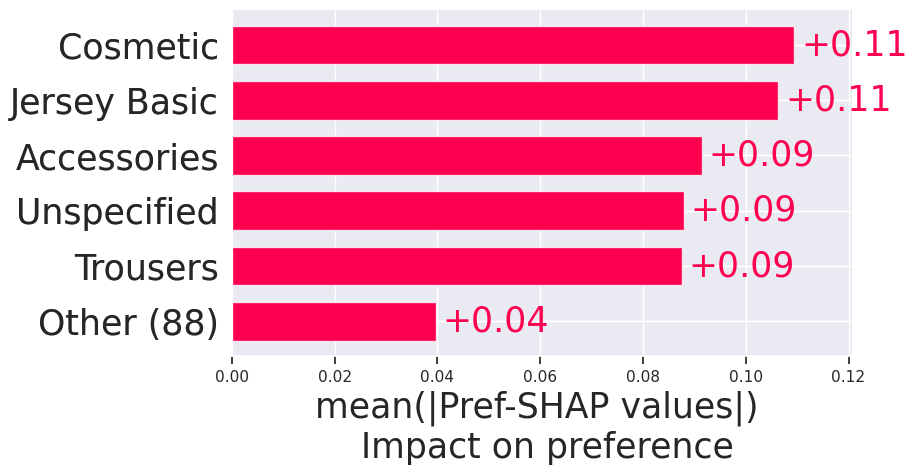}%
                \includegraphics[width=0.5\linewidth]{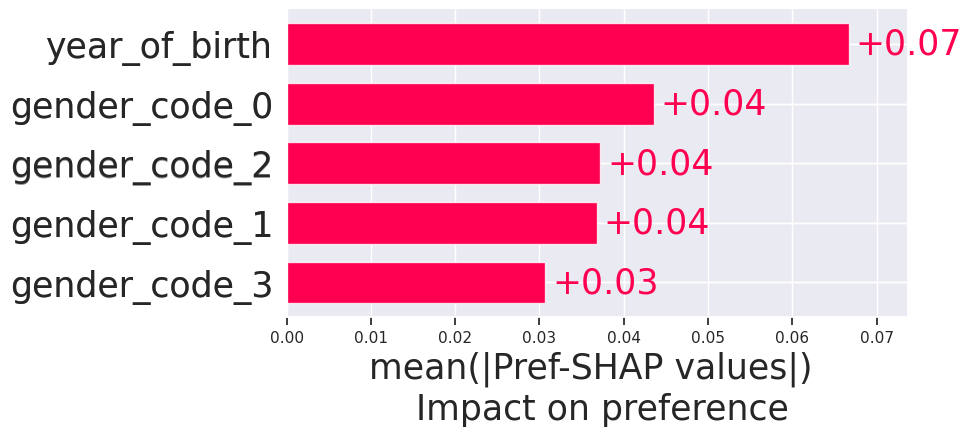}

    \end{subfigure}%
    \hfill
    \begin{subfigure}{0.495\textwidth}
        \centering
        \includegraphics[width=0.5\linewidth]{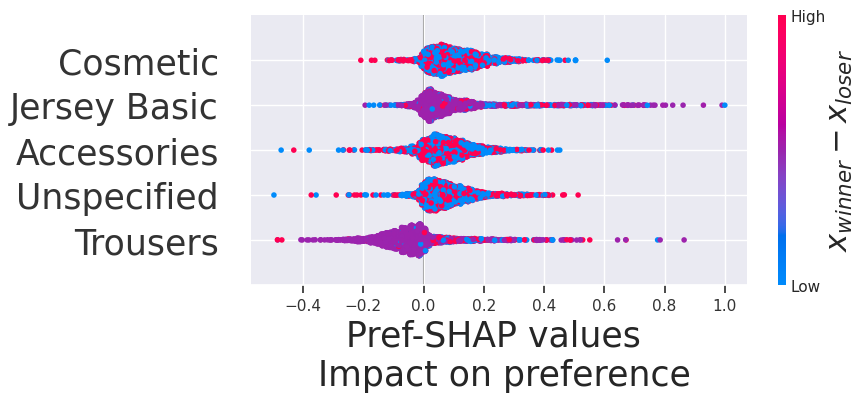}%
                \includegraphics[width=0.5\linewidth]{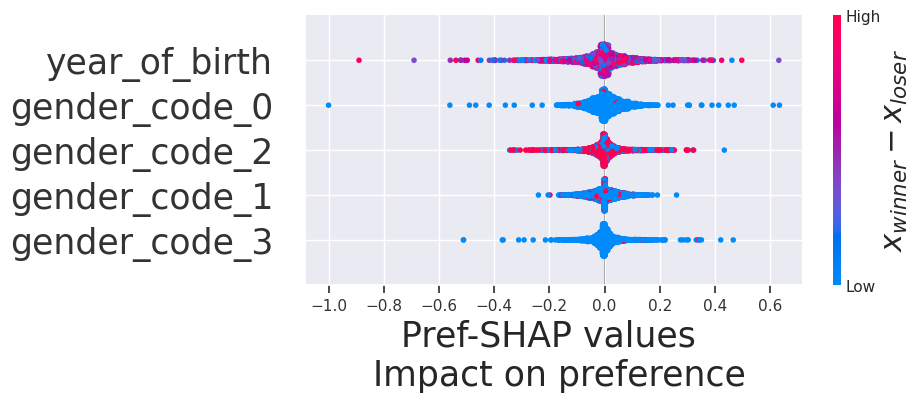}

    \end{subfigure}
            \caption{Barplots and beeplots for the website dataset, products on the left and user variables on the right. \\ \quad}
        \label{website_plots}
    \vspace{-0.5cm}
\end{figure}

\textbf{Explaining Website} \quad For the website dataset, we explain product and user preferences in \Cref{website_plots}. We generally found that, for the period considered, cosmetic products and the ``Jersey Basic category'' drove clicks.

\begin{table}[H]
\centering
 \caption{GPM vs UPM. Mean and standard deviations of performance averaged over 5 runs.}
 \label{GPL_vs_UPL}
 \resizebox{\linewidth}{!}{%

\begin{tabular}{l|cl|cl|cl|cl|cl}
\hline
         & \multicolumn{2}{c|}{Synthetic}                                 & \multicolumn{2}{c|}{Chameleon}                                  & \multicolumn{2}{c|}{Pokémon}                                    & \multicolumn{2}{c|}{Tennis}                                     & \multicolumn{2}{c}{Website}                                    \\
         & GPM                                 & \multicolumn{1}{c|}{UPM} & GPM                                  & \multicolumn{1}{c|}{UPM} & GPM                                  & \multicolumn{1}{c|}{UPM} & C-GPM                                  & \multicolumn{1}{c|}{UPM} & C-GPM                                  & \multicolumn{1}{c}{UPM} \\ \hline
Test AUC & \multicolumn{1}{l}{$0.98_{\pm0.00}$} & $0.71_{\pm0.01}$         & \multicolumn{1}{l}{$0.92_{\pm0.07}$} & $0.80_{\pm0.07}$         & \multicolumn{1}{l}{$0.86_{\pm0.00}$} & $0.82_{\pm0.00}$         & \multicolumn{1}{l}{$0.58_{\pm0.02}$} & $0.52_{\pm0.02}$         & \multicolumn{1}{l}{$0.66_{\pm0.01}$} & $0.65_{\pm0.01}$        \\
SpecR    & \multicolumn{2}{c|}{$0.09$}                                    & \multicolumn{2}{c|}{$0.24$}                                    & \multicolumn{2}{c|}{$0.20$}                                      & \multicolumn{2}{c|}{$0.13_{\pm 0.07}$}                          & \multicolumn{2}{c}{$0.53_{\pm 0.10}$}                             \\ \hline
\end{tabular}
}
\vspace{-0.5cm}
\end{table}

\begin{table}[t!]
    \centering
        \caption{Dataset summary}
    \label{dat_summary}
    \resizebox{\linewidth}{!}{%

\begin{tabular}{cccccccc} 
\hline
          Dataset & $N_{\text{Matches}}$ & $N_{\text{Players}}$ & $N_{\text{Context}}$ & $D_{\text{continuous}}$ & $D_{\text{binary}}$ & $D^{\text{Context}}_{\text{continuous}}$ & $D^{\text{Context}}_{\text{binary}}$ \\ \hline
\textit{Website}   & $85144$                & $20626$                & $129117$~(users)        & $0$                       & $93$                  & $1$                                        & $4$                                    \\ \hline
\end{tabular}
}
\vspace{-0.4cm}
\end{table}

\section{Proofs}

\newtheorem{innercustomgeneric}{\customgenericname}
\providecommand{\customgenericname}{}
\newcommand{\newcustomtheorem}[2]{%
  \newenvironment{#1}[1]
  {%
   \renewcommand\customgenericname{#2}%
   \renewcommand\theinnercustomgeneric{##1}%
   \innercustomgeneric
  }
  {\endinnercustomgeneric}
}

\newcustomtheorem{customthm}{Theorem}
\newcustomtheorem{customprop}{Proposition}
\newcustomtheorem{customlemma}{Lemma}

\begin{customprop}{3.1}[Preferential value functional for items]
Let $k$ be a product kernel on $\cX$, i.e. $k(\bfxleft,\bfxright) = \prod_{j=1}^d k^{(j)}(x^{(j)}, x'^{(j)})$. Assume $k^{(j)}$ are bounded for all $j$, then the Riesz representation of the functional $\nu^{(p)}_{\bfxleft, \bfxright, S}$ exists and takes the form:
\begin{align*}
    \nu^{(p)}_{\bfxleft, \bfxright, S} = \frac{1}{\sqrt{2}} \left(\cK(\bfxleft, S) \otimes \cK(\bfxright, S) - \cK(\bfxright, S) \otimes \cK(\bfxleft, S)\right)
\end{align*}
where $\cK(\bfx, S) = k_{S}(\cdot, \bfx_S)\otimes \mu_{X_{S^c}\mid X_S = \bfx_S}$ and $k_S(\cdot, \bfx_S) = \bigotimes_{j\in S}k^{(j)}(\cdot, x^{(j)})$ is the sub-product kernel defined analogously as $X_S$.
\end{customprop}

\begin{proof}
From ~\cite{chau2020learning}, we know the generalised preferential kernel has the following feature map:
\begin{align}
    k_E((\bfxleft,\bfxright), \cdot) = \frac{1}{\sqrt{2}}\big(k(\cdot, \bfxleft) \otimes k(\cdot, \bfxright) - k(\cdot, \bfxright)\otimes k(\cdot, \bfxleft) \big)
\end{align}
where $\otimes$ are the usual tensor product. Recall we defined the preferential value function for items as,
\begin{align}
    \nu^{(p_I)}_{\bfxleft,\bfxright,S}(g) = \EE[g(X^{(\ell)}, X^{(r)}) \mid X_S^{(\ell)} = \bfxleft_S, X_S^{(r)} = \bfxright_S]
\end{align}
as $\nu^{(p_I)}_{\bfxleft,\bfxright,S}$ is a bounded linear functional on $g$ where $g \in \cH_{k_E}$ is bounded, Riesz representation theorem~\cite{paulsen2016introduction} tells us there exists a Riesz representation of the functional in $\cH_{k_E}$, which for notation simplicity, we will denote it as $\nu^{(p_I)}_{\bfxleft,\bfxright,S}$ as well. This corresponds to,
\begin{align}
    \nu^{(p_I)}_{\bfxleft,\bfxright,S}(g) &= \EE[g(X^{(\ell)}, X^{(r)}) \mid X_S^{(\ell)} = \bfxleft_S, X_S^{(r)} = \bfxright_S] \\
    &= \langle g, \EE[k_E\left((X^{(\ell)}, X^{(r)}), \cdot \right) \mid X_S^{(\ell)} = \bfxleft_S, X_S^{(r)} = \bfxright_S] \rangle_{\cH_{k_E}} \\
    &= \langle g, \nu^{(p_I)}_{\bfxleft,\bfxright,S} \rangle_{\cH_{k_E}}
\end{align}
now we expand the expectation of the feature map as,
\begin{align}
     \nu^{(p_I)}_{\bfxleft,\bfxright,S} &= \EE\left[\frac{1}{\sqrt{2}}\big(k(\cdot, \Xleft) \otimes k(\cdot, \Xright) - k(\cdot, \Xright)\otimes k(\cdot, \Xleft) \mid X_S^{(\ell)} = \bfxleft_S, X_S^{(r)} = \bfxright_S\right]
     \label{eq: expectation_feature_map}
\end{align}
However, we note that 
\begin{align*}
    \EE[k(\cdot ,\Xleft) \otimes k(\cdot, \Xright)  \mid X_S^{(\ell)} = \bfxleft_S, X_S^{(r)} = \bfxright_S] &= \EE[k(\cdot, X)\mid X_S = \bfxleft_S] \\ 
    &\quad \quad \quad \otimes \EE[k(\cdot, X)\mid X_S=\bfxright_S],
\end{align*}
because $\Xleft$ and $\Xright$ are identical copies of $X$ and we take the reference distribution as $p(\Xleft, \Xright \mid X_S^{(\ell)} = \bfxleft_S, X_S^{(r)} = \bfxright_S) = p(\Xleft\mid \Xleft_S = \bfxleft_S)p(\Xright \mid \Xright_S=\bfxright_S)$. Focusing on the duplicating component, we have,
\begin{align}
    \EE[k(\cdot, X)\mid X_S = \bfxleft] &= \EE[k_S(\cdot, X_S) \otimes k_\Sc(\cdot, X_\Sc) \mid X_S=\bfxleft_S] \\
    &= k_S(\cdot, \bfxleft_S) \otimes \EE[k_\Sc(\cdot, X_\Sc)\mid X_S=\bfxleft_S] \\
    &= k_S(\cdot, \bfxleft_S) \otimes \mu_{X_\Sc\mid X_S=\bfxleft_S}\\
    &=: \cK(\bfxleft, S)
\end{align}
therefore by symmetry, we can arrange the terms in Eq~\ref{eq: expectation_feature_map} and conclude the proposition,
\begin{align}
    \nu^{(p_I)}_{\bfxleft, \bfxright, S} = \frac{1}{\sqrt{2}} \left(\cK(\bfxleft, S) \otimes \cK(\bfxright, S) - \cK(\bfxright, S) \otimes \cK(\bfxleft, S)\right)
\end{align}
\end{proof}
To estimate the preferential value functional, we simply replace the conditional mean embeddings with the empirical versions, i.e. $\hat{\cK}(\bfx, S) = k_S(\cdot, \bfx_S) \otimes \hat{\mu}_{X_{\Sc}\mid X_S=\bfx_S}$, where $\hat{\mu}_{X_{\Sc}\mid X_S=\bfx_S} = \bfK_{\bfx_S, \bfX_S}(\bfK_{\bfX_S,\bfX_S} + n\lambda I)^{-1}{\bf \Phi}_{X_{\Sc}}^\top$ is the standard conditional mean embedding estimator (${\bf \Phi}_{X_{\Sc}}$ is the feature map matrix of rv $X_{\Sc}$).

Now we proceed to estimate the preferential value function given a function $g$ from the RKHS,
\begin{customprop}{3.2}[Non-parametric Estimation]
Given $\hat{g} = \sum_{j=1}^m \alpha_j k_E((\bfxleft_j, \bfxright_j), \cdot)$, datasets $\bfXleft, \bfXright$, test items $\bfxleft, \bfxright$, the preferential value function at test items $\bfxleft, \bfxright$ for coalition $S$ and preference function $\hat{g}$ can be estimated as
\begin{align*}
    \hat{\nu}_{\small \bfxleft,\bfxright, S}^{(p_I)}(\hat{g}) = \balpha^\top\left(\Gamma(\bfXleft_S, \bfxleft_S)\odot \Gamma(\bfXright_S, \bfxright_S) - \Gamma(\bfXleft_S, \bfxright_S)\odot \Gamma(\bfXright_S, \bfxleft_S)\right),
\end{align*}
where $\Gamma(\bfXleft_S, \bfxleft_S) = \bfK_{\bfXleft_S, \bfxleft_S}\odot \bfK_{\bfXleft_{S^c}, \bfX_{S^c}}\bfK_{\bfX_S, \lambda}^{-1}\bfK_{\bfX_S, \bfxleft_S}, \bfK_{\bfX_S, \lambda} = \bfK_{\bfX_S, \bfX_S} + n\lambda I$, $\balpha = \{\alpha_j\}_{j=1}^m$ and $\lambda > 0$ is a regularisation parameter.
\end{customprop}

\begin{proof}
Given $\hat{g}$, the preferential value function evaluated at $\hat{g}$ can be written as,
\begin{align}
    \hat{\nu}_{\small \bfxleft,\bfxright, S}^{(p_I)}(\hat{g}) &= \langle\hat{g},  \hat{\nu}_{\small \bfxleft,\bfxright, S}^{(p_I)}\rangle_{\cH_{k_E}}\\
    &= \langle \sum_{j=1}^m \alpha_j k_E((\bfxleft_j, \bfxright_j), \cdot), \frac{1}{\sqrt{2}} \left(\hat{\cK}(\bfxleft, S) \otimes \hat{\cK}(\bfxright, S) - \hat{\cK}(\bfxright, S) \otimes \hat{\cK}(\bfxleft, S)\right)\rangle_{\cH_{k_E}} \\
    &= \frac{1}{\sqrt{2}} \left\langle \sum_{j=1}^m \alpha_j k_E((\bfxleft_j, \bfxright_j), \cdot), \hat{\cK}(\bfxleft, S) \otimes \hat{\cK}(\bfxright, S) \right\rangle \\
    &\quad\quad\quad - \frac{1}{\sqrt{2}} \left\langle \sum_{j=1}^m \alpha_j k_E((\bfxleft_j, \bfxright_j), \cdot), \hat{\cK}(\bfxright, S) \otimes \hat{\cK}(\bfxleft, S) \right\rangle
\end{align}
Now we focus on the first component, and rewrite:
\begin{align}
    \frac{1}{\sqrt{2}} \left\langle \sum_{j=1}^m \alpha_j k_E((\bfxleft_j, \bfxright_j), \cdot), \hat{\cK}(\bfxleft, S) \otimes \hat{\cK}(\bfxright, S) \right\rangle = \sum_{j=1}^m A_j^{(1)}
\end{align}
and we continue to expand the terms,
\begin{align}
    A_j^{(1)} &:=\frac{1}{\sqrt{2}}\left\langle \alpha_j k_E((\bfxleft_j, \bfxright_j), \cdot), \hat{\cK}(\bfxleft, S) \otimes \hat{\cK}(\bfxright, S) \right\rangle \\
    &= \frac{1}{\sqrt{2}}\left\langle\frac{\alpha_j}{\sqrt{2}}\big(k(\cdot, \bfxleft_j) \otimes k(\cdot, \bfxright_j) - k(\cdot, \bfxright_j)\otimes k(\cdot, \bfxleft_j) \big), \hat{\cK}(\bfxleft, S) \otimes \hat{\cK}(\bfxright, S) \right\rangle \\
    &= \frac{\alpha_j}{2}\left(\left\langle k(\cdot, \bfxleft_j) \otimes k(\cdot, \bfxright_j), \hat{\cK}(\bfxleft, S) \otimes \hat{\cK}(\bfxright, S) \right\rangle - \left\langle k(\cdot, \bfxright_j)\otimes k(\cdot, \bfxleft_j) , \hat{\cK}(\bfxleft, S) \otimes \hat{\cK}(\bfxright, S)\right\rangle\right) \\
    &=\frac{\alpha_j}{2} \left(A_j^{(1, \ell)} - A_j^{(1, r)} \right)
\end{align}
We then note that 
\begin{align}
    A_j^{(1, \ell)}&:= \left\langle k(\cdot, \bfxleft_j)) \otimes k(\cdot, \bfxright_j), \hat{\cK}(\bfxleft, S) \otimes \hat{\cK}(\bfxright, S) \right\rangle \\ &= \left\langle k(\cdot, \bfxleft_j), \hat{\cK}(\bfxleft, S) \right\rangle\left\langle k(\cdot, \bfxright_j), \hat{\cK}(\bfxright, S) \right\rangle \\
    &= k_S(\bfxleft_j, \bfxleft)\bfK_{{\bfxleft_j}_S, \bfX_S}(\bfK_{\bfX_S,\bfX_S} + n\lambda I)^{-1}\bfK_{\bfX_{\Sc}, \bfxleft_{\Sc}} \\
    &\quad\quad\quad \times k_S(\bfxright_j, \bfxright)\bfK_{{\bfxright_j}_S, \bfX_S}(\bfK_{\bfX_S,\bfX_S} + n\lambda I)^{-1}\bfK_{\bfX_{\Sc}, \bfxright_{\Sc}} \\
    &= \Gamma(\bfxleft_{j_S}, \bfxleft_S) \odot \Gamma(\bfxright_{j_S}, \bfxright_S)
\end{align}
To go from the second equation to the third equation in this paragraph, realise $k(\cdot, \bfxleft) = k_S(\cdot, \bfxleft_S)\otimes k_{\Sc}(\cdot, \bfxleft_{\Sc})$ by product kernel assumption. 
In this case, we can rewrite $A_j^{(1)}$ as,
\begin{align}
    A_j^{(1)} = \frac{\alpha_j}{2}\left(\Gamma(\bfxleft_{j_S}, \bfxleft_S) \odot \Gamma(\bfxright_{j_S}, \bfxright_S) - \Gamma(\bfxright_{j_S}, \bfxleft_{S}) \odot \Gamma(\bfxleft_{j_S}, \bfxright_S)\right)
\end{align}
Analogously, define $\sum A_j^{(2)}$ as the second component after the subtraction sign, by symmetry, we know
\begin{align}
    A_j^{(2)} = \frac{\alpha_j}{2}\left(\Gamma(\bfxleft_{j_S}, \bfxright_S)\odot \Gamma(\bfxright_{j_S}, \bfxleft_S) - \Gamma(\bfxright_{j_S}, \bfxright_{S}) \odot \Gamma(\bfxleft_{j_S}, \bfxleft_S)\right)
\end{align}
by subtracting $A_j^{(1)}$ and $A_j^{(1)}$, we get the following:
\begin{align}
    \hat{\nu}_{\small \bfxleft,\bfxright, S}^{(p_I)}(\hat{g}) &= \sum_{j=1}^m \alpha_j \left(\Gamma(\bfxleft_{j_S}, \bfxleft_S) \odot \Gamma(\bfxright_{j_S}, \bfxright_S) - \Gamma(\bfxright_{j_S}, \bfxleft_S) \odot \Gamma(\bfxleft_{j_S}, \bfxright_S)\right)\\
    \intertext{writing it in compact form, we arrive to our result,}
    &= \balpha^\top\left(\Gamma(\bfXleft_S, \bfxleft_S)\odot \Gamma(\bfXright_S, \bfxright_S) - \Gamma(\bfXleft_S, \bfxright_S)\odot \Gamma(\bfXright_S, \bfxleft_S)\right)
\end{align}
\end{proof}

\begin{customprop}{3.3}[Preferential value function for contexts] Given a preference function $g_U\in \cH_{k_E^U}$, denote $\Omega' = \{1,...,d'\}$, then the utility of context features $S'\subseteq \Omega'$ on $\{\bfu, \bfxleft,\bfxright\}$ is measured by $\nu^{(p_U)}_{\bfu, \bfxleft,\bfxright, S'}(g_U) = \EE[g_U\left(\{\bfu_{S'}, U_{{S'}^c}\}, \bfxleft,\bfxright\right)\mid U_{S'}=\bfu_{S'}]$ where the expectation is taken over the observational distribution of $U$. Now, given a test triplet ($\bfu, \bfxleft, \bfxright)$, if $\hat{g}_U= \sum_{j=1}^m \alpha_j k_E^U\left((\bfu_j, \bfxleft_j, \bfxright_j), \cdot\right)$, the non-parametric estimator is:
\begin{align*}
    \hat{\nu}^{(p_U)}_{\bfu,\bfxleft,\bfxright, S'}(\hat{g}_U) &= \balpha^\top \left(\left( \bfK_{\bfU_{S'}, \bfu_{S'}} \odot \bfK_{\bfU_{S'^c}, \bfU_{S'^c}}\left(\bfK_{\bfU_{S'}, \bfU_{S'}} + m\lambda' I\right)^{-1}\bfK_{\bfU_{S'}, \bfu_{S'}}\right)\odot \Xi_{\bfxleft,\bfxright} \right)
\end{align*}
where $\Xi_{\bfxleft,\bfxright} = \left(\bfK_{\bfXleft, \bfxleft}\odot \bfK_{\bfXright, \bfxright} - \bfK_{\bfXright, \bfxleft}\odot \bfK_{\bfXleft, \bfxright}\right)$.
\end{customprop}
\begin{proof}
Recall the feature map of the kernel $k_E^U$ takes the following form,
\begin{align}
    k_E^U\left((\bfu, \bfxleft,\bfxright), \cdot\right) = k_u(\bfu, \cdot) \otimes k_E((\bfxleft, \bfxright), \cdot)
\end{align}
Therefore we can express the preferential value function for context as,
\begin{align}
    \nu^{(p_U)}_{\bfu, \bfxleft,\bfxright, S'}(g_U) &= \EE\left[g_U\left(\{\bfu_{S'}, U_{\Sc}\}, \bfxleft,\bfxright\right)\mid U_S=\bfu_{S'}\right] \\
    &= \left\langle g_U, \EE\left[k_E^U\left((\{\bfu_{S'}, U_{{S'}^c}\}, \bfxleft,\bfxright), \cdot \right) \mid U_{S'} = \bfu_{S'}\right]\right\rangle \\
    &= \left\langle g_U, \EE\left[k_u(\{\bfu_{S'}, U_\Sc\} \mid U_{S'} = \bfu_{S'}\right] \otimes k_E((\bfxleft, \bfxright), \cdot)\right\rangle \\
    &=  \left\langle g_U, k_{u_{S'}}(\bfu_{S'})\otimes \mu_{U_{{S'}^c}\mid U_{S'}=\bfu_{S'}} \otimes k_E((\bfxleft, \bfxright), \cdot)\right\rangle
\end{align}
The remaining steps are analogous to~\cite[Prop.2]{chau2021rkhs}. To obtain the empirical estimation, we first replace the conditional mean embedding $\mu_{U_{{S'}^c} \mid U_{S'}=\bfu_{S'}}$ with its empirical estimate and replace $g_U$ with $\hat{g}_U = \sum_{j=1}^m \alpha_j k_E^U\left((\bfu_j, \bfxleft_j, \bfxright_j), \cdot\right)$. Now the empirical estimator has the following form,
\begin{align}
\small
    \hat{\nu}^{(p_U)}_{\bfu,\bfxleft,\bfxright, S'}(\hat{g}_U) &= \left\langle \sum_{j=1}^m \alpha_j k_E^U\left((\bfu_j, \bfxleft_j, \bfxright_j), \cdot \right) , k_{u_{S'}}(\bfu_{S'})\otimes \hat{\mu}_{U_{{S'}^c}\mid U_{S'}=\bfu_{S'}} \otimes k_E((\bfxleft, \bfxright), \cdot)\right\rangle \\
    &=\sum_{j=1}^m \alpha_j \left\langle k_u(\bfu_j, \cdot) \otimes k_E((\bfxleft_j, \bfxright_j), \cdot) , k_{u_{S'}}(\bfu_{S'})\otimes \hat{\mu}_{U_{{S'}^c}\mid U_{S'}=\bfu_{S'}} \otimes k_E((\bfxleft, \bfxright), \cdot)\right\rangle \\
    &= \sum_{j=1}^m \alpha_j\left\langle k_u(\bfu_j, \cdot), k_{u_S}(\bfu_S) \otimes \hat{\mu}_{U_{{S'}^c}\mid U_S=\bfu_S}\right\rangle k_E\left((\bfxleft_j, \bfxright_j), (\bfxleft, \bfxright)\right)\\
\intertext{Now write everything in terms of matrices,}
&= \balpha^\top \left(\left( \bfK_{\bfU_{S'}, \bfu_{S'}} \odot \bfK_{\bfU_{S'^c}, \bfU_{S'^c}}\left(\bfK_{\bfU_{S'}, \bfU_{S'}} + m\lambda' I\right)^{-1}\bfK_{\bfU_{S'}, \bfu_{S'}}\right)\odot \Xi_{\bfxleft,\bfxright} \right)
\end{align}
where $\Xi_{\bfxleft,\bfxright} = \left(\bfK_{\bfXleft, \bfxleft}\odot \bfK_{\bfXright, \bfxright} - \bfK_{\bfXright, \bfxleft}\odot \bfK_{\bfXleft, \bfxright}\right)$.
\end{proof}

\end{document}